\documentclass[10pt,twocolumn,letterpaper]{article}

\usepackage[pagenumbers]{iccv} % To force page numbers, e.g. for an arXiv version

\usepackage[pagebackref,breaklinks,colorlinks,allcolors=blue]{hyperref}
\usepackage{multirow}
\usepackage{xcolor}
\usepackage{amsfonts}
\usepackage{pifont}
\usepackage{enumitem}
\usepackage{algorithm}   
\usepackage{algorithmic} 
\def\paperID{*****} % *** Enter the Paper ID here
\def\confName{ICCV}
\def\confYear{2025}

\title{Enhancing Visual Reliance in Text Generation: 
A Bayesian Perspective on Mitigating Hallucination in Large Vision-Language Models}

\author{Nanxing Hu\textsuperscript{1,2}, Xiaoyue Duan\textsuperscript{2}, Jinchao Zhang\textsuperscript{2},  Guoliang Kang\textsuperscript{1}\thanks{Corresponding author}\\
\textsuperscript{1}Beihang University; \textsuperscript{2}Tencent WXG \\
\tt neilhnx@buaa.edu.cn
}

\begin{document}

\maketitle

\begin{abstract}
Large Vision-Language Models (LVLMs) usually generate texts which satisfy context coherence but don't match the visual input. 
Such a hallucination issue hinders LVLMs' applicability in the real world.
The key to solving hallucination in LVLM is to make the text generation rely more on the visual content. 
Most previous works choose to enhance/adjust the features/output of a specific modality (i.e., visual or textual) to alleviate hallucinations in LVLM, which do not explicitly or systematically enhance the visual reliance.
In this paper, we comprehensively investigate the factors that may degenerate the visual reliance in text generation of LVLM from a Bayesian perspective. We propose to mitigate hallucination in LVLM from three aspects. 
Firstly, we observe that not all visual tokens are informative in generating meaningful texts. We propose to evaluate and remove redundant visual tokens to avoid their disturbance. 
Secondly, LVLM may encode inappropriate prior information, making it lean toward generating unexpected words.
We propose a simple, yet effective way to rectify the prior from a Bayesian perspective.
Thirdly, we observe that starting from certain steps, the posterior of next-token prediction conditioned on visual tokens may collapse to a prior distribution which does not depend on any informative visual tokens at all. 
Thus, we propose to stop further text generation to avoid hallucination.
Extensive experiments on three benchmarks, including POPE, CHAIR, and MME, demonstrate that our method can consistently mitigate the hallucination issue of LVLM and performs favorably against previous state-of-the-arts. Codes are available at \url{https://github.com/NeilHnxTcc/EVRB}.
\end{abstract}

\begin{figure*}[t]
  \centering
  \includegraphics[width=\linewidth]{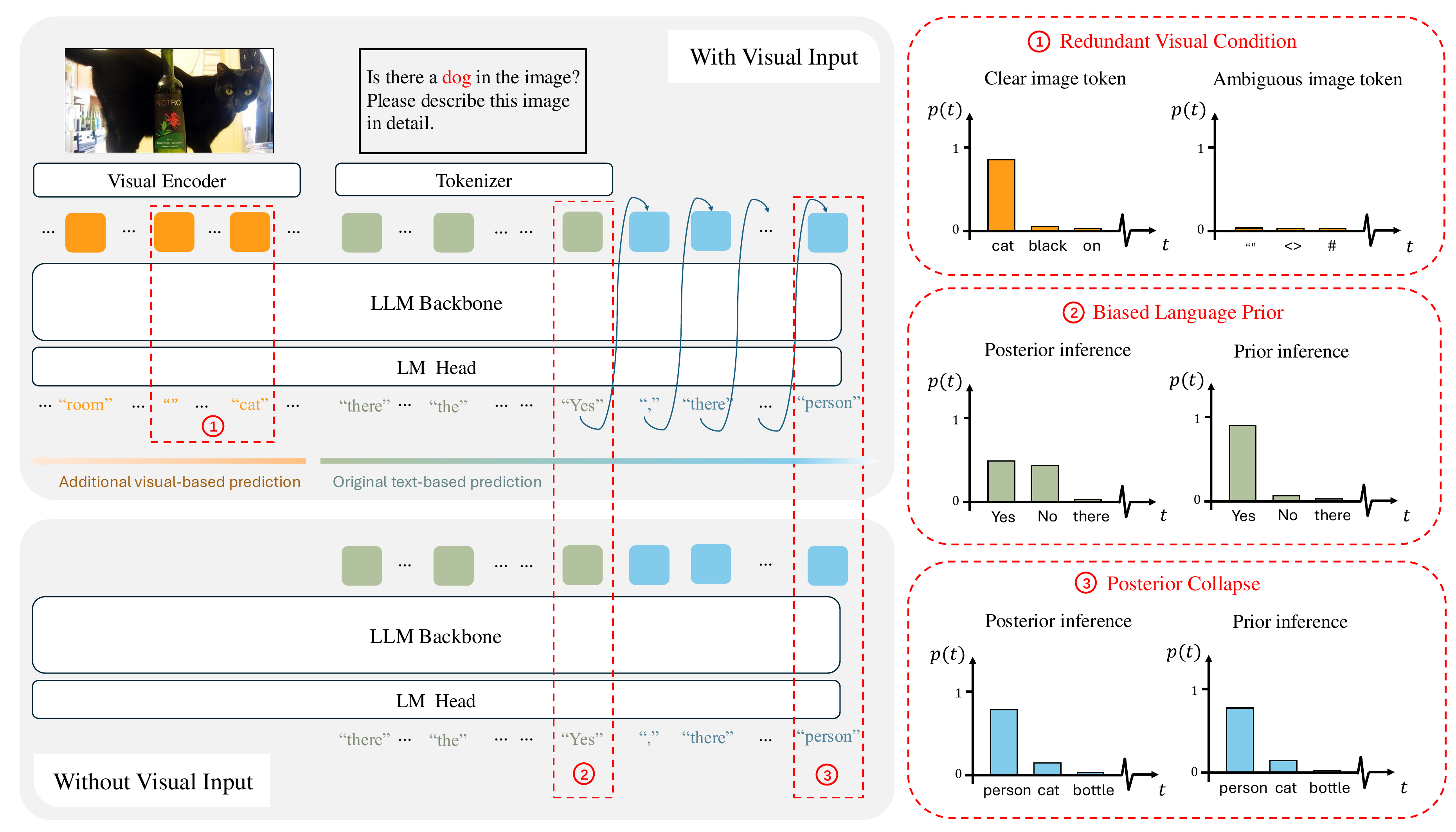} % 调整 width 控制图片宽度
  \caption{Hallucination causes from a Bayesian view. The posterior inference process of LVLM (left-up) is conditioned on both visual and textual information, while the prior process (left-down) is only conditioned on textual one. Three issues that can incur hallucinations (framed by red boxes). \ding{192} Redundant visual condition: Compared to clear visual tokens that carry explicit visual information, ambiguous visual tokens serve as redundant conditions that disturb the inference. \ding{193} Biased language prior: The inherent bias of language prior can lead to posterior inference, making incorrect predictions. \ding{194} Posterior collapse: Posterior inference makes a similar prediction as prior in the late stage of generation, failing to maintain visual-textual consistency.} 
  \label{fig:intro}
\end{figure*}

\section{Introduction}
\label{sec:introduction}
Large Vision-Language Models (LVLMs) \cite{liang2024survey,wang2024qwen2,achiam2023gpt,yao2024minicpm,liu2023visual,liu2024improved}, designed to integrate and process both visual and textual information, demonstrate remarkable multi-modal understanding and reasoning capabilities.
However, in practice, LVLMs suffer from a critical issue known as hallucination: LVLMs generate context-coherent content while not all of them exist in the visual input \cite{li2023evaluating}. 
This problem poses serious risks in the applications of LVLMs \cite{zhang2023huatuogpt,wang2023chatcad,hu2023advancing,cui2024survey,wang2023drivemlm,brie2023evaluating},
where accurate and reliable reasoning is required.

The fundamental solution for alleviating hallucination in LVLMs lies in enhancing the visual reliance in text generation. 
Most previous works choose to enhance the features or adjust the outputs of a specific modality \cite{huo2024self,leng2024mitigating,wang2024mitigating,huang2024opera} (\emph{i.e.,} visual or textual) to alleviate hallucinations in LVLM. However, they ignore explicitly enhancing the reliance of text generation on visual inputs, which largely constrains the hallucination mitigation effect. 
%do not explicitly or systematically enhance the visual reliance.
%To achieve this, both modalities and their interactive mechanism should be taken into account. 
%while current works tend to attribute the cause of hallucination to some isolated issues, choose to enhance the features or adjust the output from a specific modal perspective \cite{huo2024self,leng2024mitigating,wang2024mitigating,huang2024opera}. Their designs do not explicitly and systematically enhance the visual reliance, resulting in an incomplete resolution to the hallucination. 

% Compared to the hallucination of unimodal LLMs \cite{chen2024context,ji2023survey,lu2024insights}, the reasons for hallucination in LVLMs are more complex. We need to take all the modal and their interactive mechanism into account to analyze the cross-modality inconsistency. while current works tend to attribute the cause of hallucination to some isolated issues, such as visual bias \cite{huo2024self,leng2024mitigating} or language bias \cite{wang2024mitigating,huang2024opera}. They lack a comprehensive investigation into the underlying mechanism, leading to an incomplete resolution to the hallucination. 

In this paper, we comprehensively investigate the factors that may degenerate the visual reliance in text generation of LVLM. 
The text generation of LVLM can be modeled as a conditional probability prediction, \emph{i.e.,} the probability of the next text token prediction should be conditioned on the visual tokens and previous text tokens.
From a Bayesian perspective, we comprehensively investigate three aspects that may harm the visual-conditioned text generation and correspondingly propose three modules to enhance the reliance of text on visual information to mitigate the hallucinations in LVLM.

Firstly, we observe that quite a few visual tokens carry ambiguous and meaningless information, serving as redundant conditions in Bayesian inference. 
In practice, these redundant visual tokens may increase the uncertainty of text predictions and impair the modeling of text's reliance on informative visual information.  
%Compared to text tokens, visual tokens contribute a large percentage of input tokens while capturing lower information density. 
As illustrated in Figure \ref{fig:intro}, problem \ding{192}, we perform additional next token predictions based on visual tokens to evaluate their quality. 
Some image tokens can make a confident prediction aligned with the image content (see the distribution of next token prediction based on ``clear image token'') while others make an ambiguous and meaningless prediction (see ``ambiguous image token'').

Secondly, we find that the language prior inherits the bias from LLM, which confuses the text generation based on visual inputs. 
Given a visual-tasks-related instruction, an ideal prior distribution, \emph{i.e.,} the probability of the next text token purely conditioned on previous texts, should approximate the uniform distribution over a subset vocabulary given that it doesn’t have any visual information. 
However, the language bias exists, which means the LVLM inherently tends to assign higher probabilities to certain words even without any informative visual information.
%However, the language prior inherits the bias from LLM, which can undermine the Bayesian inference. 
As illustrated in Figure \ref{fig:intro}, problem \ding{193}, the prior inference makes a confident but incorrect answer to the object existence question. 
%Affected by the inappropriate biased prior, the posterior inference is distorted and results in hallucination.

%To rectify the language prior, we propose a simple, yet effective way from a Bayesian perspective. 
%Given that the posterior is proportional to the prior and the ideal prior should be a uniform distribution over a vocabulary subset, we can remove the prior bias by dividing the posterior probability directly by the prior probability. % on a truncated vocabulary.

Thirdly, we observe that the posterior distribution conditioned on visual inputs will collapse into the prior distribution, which does not rely on any informative visual tokens at all as the length of the generated text increases. 
As illustrated in Figure \ref{fig:intro}, problem \ding{194}, the current next token prediction is about the object presented in the image. The distribution of posterior and prior inference should be different because one has visual information while the other can only guess according to the text context.
However, the posterior distribution is similar to the prior distribution, indicating that no visual information is used to make this prediction.
Under this circumstance, the visual reliance is no longer guaranteed, and the model outputs a hallucination word, \textit{``person''}.

Based on our observations, we propose a training-free framework named EVRB (Enhancing Visual Reliance from a Bayesian perspective) to enhance the visual reliance in text generation and thus mitigate the hallucination issue of LVLM. Firstly, to reduce the negative effect of non-informative (or redundant) visual tokens on text token prediction, we propose to evaluate the image tokens based on their next token prediction and remove the redundant ones. 
Specifically, we leverage the entropy of the predicted probability to judge the quality of the visual tokens.
The tokens with high entropy are treated as ambiguous and meaningless, and we remove them to enhance the informative visual reliance in text generation.
Secondly, to rectify the language prior, we propose a simple yet effective way from a Bayesian perspective. 
Given that the posterior is proportional to the prior and the ideal prior should be a uniform distribution over a vocabulary subset, we can remove the prior bias by dividing the posterior probability directly by the prior probability. 
Thirdly, we choose to stop the inference process to avoid hallucination before the posterior collapses into the prior. Specifically, we first locate the ``factual'' text tokens (the prediction of which should be conditioned on informative visual inputs) based on the value-value attention map. %to eliminate the interference of the tokens that naturally only rely on the textual content. 
We monitor the distance (JS divergence) between the posterior of these factual text tokens and the prior and utilize the distance evaluation to modify the occurring probability of the termination token.
In this way, we expect the model can terminate the text generation in time if the posterior distribution collapses into the prior distribution.
Experiments on three benchmarks including POPE~\cite{li2023evaluating}, CHAIR~\cite{rohrbach2018object}, and MME~\cite{fu2306mme} verify the effectiveness of our proposed framework. Results show that our method performs favorably against previous state-of-the-arts.

In a nutshell, our contributions are summarized as follows
\begin{itemize}[leftmargin=*]

    \item  We observe three core factors undermining visual reliance in LVLMs from a Bayesian perspective: ambiguous visual tokens serving as redundant conditions, biased language priors undermining the posterior inference, and posterior distribution collapsing into prior during generation.

   \item We propose a training-free framework EVRB, to address these issues: pruning the ambiguous visual tokens with high entropy to get rid of the redundant condition; rectifying the biased language prior by dividing the posterior probability directly by the prior probability, and applying appropriate early stopping based on the JS divergence to prevent posterior collapse. 

    \item Extensive experiments on three benchmarks, including POPE~\cite{li2023evaluating}, CHAIR~\cite{rohrbach2018object}, and MME~\cite{fu2306mme}, show our framework consistently mitigates the hallucination issue of LVLM. Results show that our method performs favorably against previous state-of-the-arts.
    
\end{itemize}

\section{Related Work}
\label{sec:related_work}
\textbf{Large Vision-Language Models.}
In recent years, Large Language Models (LLMs) have achieved significant advancements \cite{raffel2020exploring,brown2020language,chowdhery2023palm,taori2023stanford,gilardi2023chatgpt,touvron2023llama,bai2023qwen}. 
Incorporating LLM with a visual encoder \cite{radford2021learning} and a cross-modal alignment connector, Large Vision-Language Models (LVLMs) \cite{liang2024survey,wang2024qwen2,achiam2023gpt,yao2024minicpm,liu2023visual,liu2024improved} can simultaneously deal with visual and textural information and have demonstrated remarkable capabilities across various of multimodal tasks, including image captioning \cite{hossain2019comprehensive,zhu2023minigpt} and visual question answering \cite{wang2024surgical,dang2024sadl}. 
These tasks place high demands on the accuracy and reliability of responses. However, LVLMs suffer from serious hallucination issues \cite{li2023evaluating,liu2023aligning}, which undermine their reliability and restrict their application in the real world.

\noindent \textbf{Hallucination Mitigation in LVLMs.} 
In the context of LLMs, hallucination manifests as generating context not aligned with world knowledge or nonsensical content \cite{chen2024context,ji2023survey,lu2024insights}. 
For LVLMs, the main problem is whether the generated contents are consistent with the provided visual information. Hallucination can significantly undermine LVLMs' performance \cite{zhao2023beyond,wang2023caption}. 
Various methods have been proposed to mitigate the hallucinations. Some works aim to mitigate hallucination by fine-tuning LVLMs on additional datasets \cite{yue2024less,yin2024woodpecker,sun2023aligning,liu2024survey,gunjal2024detecting}. However, the requirement for extra training and high-quality data imposes practical limitations on their applicability. 
Thus, training-free methods have attracted widespread attention. 
Some works mitigate hallucination by applying contrastive decoding strategy \cite{huo2024self,wang2024mitigating,leng2024mitigating};
some leverage the intermediate feature to assist the generation \cite{wang2024mllm,zhou2024mitigating};
others adopt well-designed inference strategies \cite{huang2024opera,yue2024less}. 
However, these methods ignore explicitly enhancing the reliance of text generation on visual inputs, which is the fundamental requirement for alleviating hallucination, and therefore can hardly achieve a satisfactory overall performance.
In contrast, we make an investigation into the factors that may degenerate the visual reliance in text generation of LVLM based on Bayesian theory, and propose a training-free strategy EVRB that can generally mitigate the hallucination. 

\section{METHODS}
\label{sec:Method}

\subsection{Preliminary}
LVLMs integrate visual and textual inputs to generate coherent and context-aware responses. We formalize their core architecture and generation mechanism as follows.

\textbf{Modality alignment.} Take advantage of pre-trained vision-language models, like CLIP \cite{radford2021learning} and a multi-modal projector, LVLMs manage to align the visual and textual latent space. 
The raw image will be mapped into visual tokens $\textbf{v}=\{ v_0, v_1, \dots, v_{n_s-1}\}$ 
and the text inputs are tokenized into text tokens $\textbf{t}=\{ t_0, t_1, \dots, t_{n_t-1}\}$. 
$n_s$ and $n_t$ are the numbers of visual and textual tokens, respectively. 
$n_t$ increases with the number of generated tokens.
The input sequence is formed by concatenating the image tokens and text tokens.

\textbf{LVLM's inference.} The input sequence is processed by the backbone networks: a pre-trained LLM consisting of $L$ transformer layers. 
The intermediate features generated by $i$-th layer are called hidden states, denoted as $\textbf{h}^i=\{h_0^i, h_1^i, \dots, h^i_{n-1}\}$, where $n=n_s+n_t$. 
The hidden state of last token in the last layer $L$ is mapped by the language model head $f(\cdot)$ to predict the probability of the next token among the vocabulary set $\Phi$:
\begin{equation}
    p(t_{n_t}|\textbf{v},\textbf{t}) = \text{softmax}(f(h^L_{n-1})), \, t_{n_t} \in \Phi
\end{equation}
where $p(t_{n_t}|\textbf{v},\textbf{t})$ denotes the probability of the next word conditioned on the input sequence. The predicted token is concatenated to the end of the input sequence for the next token generation, and the model keeps generating new tokens auto-regressively until the termination token $\text{<EOS>}$ is generated.

\subsection{Redundant Visual Token Pruning}
\label{subsec:token_pruning}
The visual tokens carrying ambiguous and meaningless information serve as redundant conditions in Bayesian inference.
They can impair visual reliance in text generation and cause hallucination.
To mitigate this problem, we need to find a metric to rate the image tokens and eliminate the redundant ones. 

The LVLM is pre-trained for multi-modal feature alignment on a large scale of data \cite{liu2024improved}. 
So the latent space of image tokens is aligned with the latent space of text tokens. 
We assume that it is valid to use image tokens for the next token prediction, and the results can reflect the visual information carried by these tokens. 
Thus, we apply the language model head $f(\cdot)$ to the hidden states of image tokens in the last layer, \emph{i.e.,}
\begin{equation}
    p(t_{v}|v_{\leq i}) = \text{softmax}(f(h^L_{i})), 
\end{equation}
where $i \in \{0,1, \dots, n_s-1\}$, and $t_v$ is the text prediction in the vocabulary set $\Phi$, purely based on the visual information.
Consistent with our assumption, the predicted probabilities of image tokens show a good alignment with the image semantic information. 

\begin{figure}[h]
  \centering
  \subfloat[]{
      \includegraphics[width=0.226\textwidth]{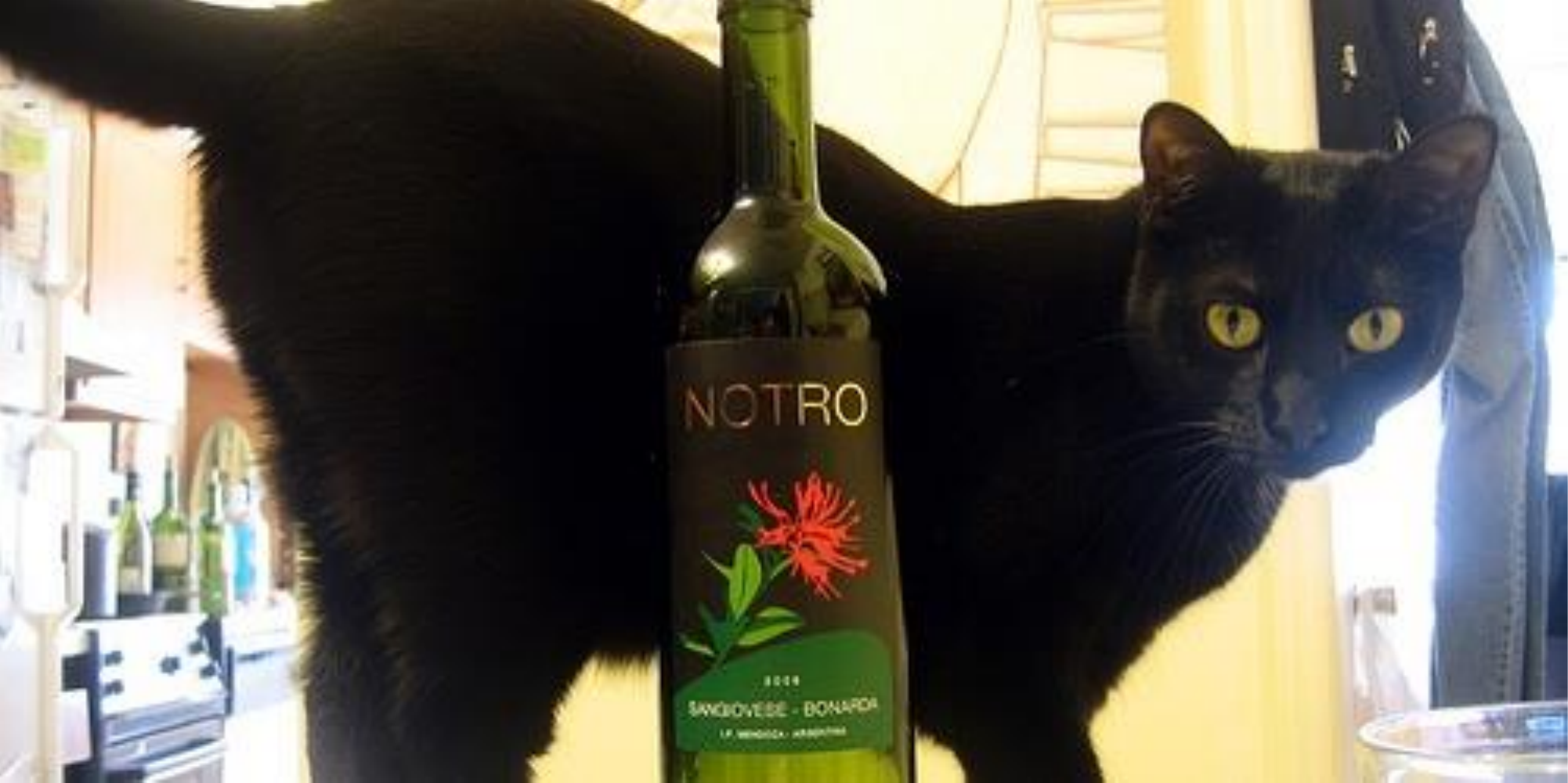}
      \label{subfig:cat_bottle}
  }
  \subfloat[]{
      \includegraphics[width=0.226\textwidth]{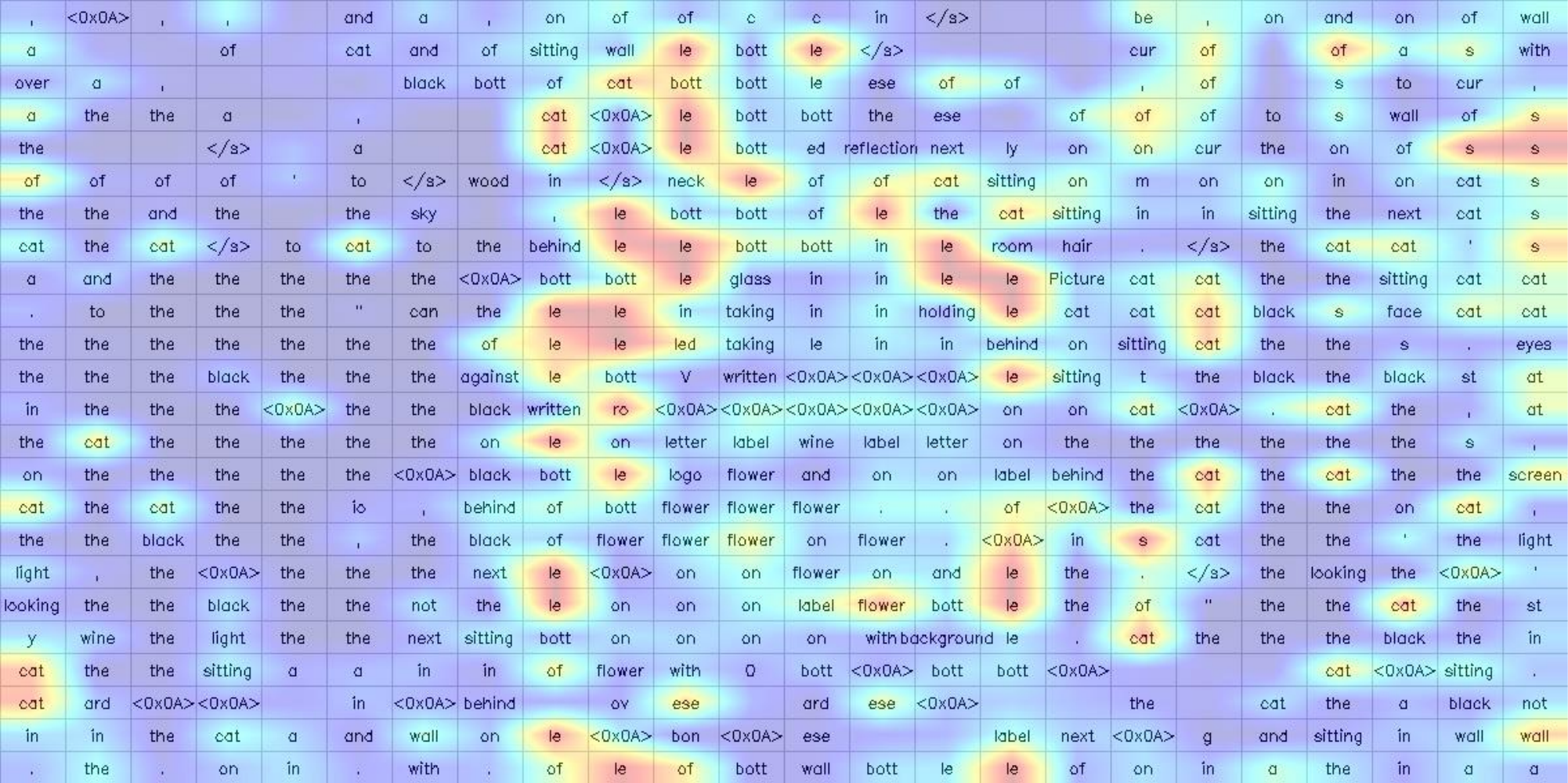}
      \label{subfig:text_entropy}
  }

  \subfloat[]{
      \includegraphics[width=0.46\textwidth]{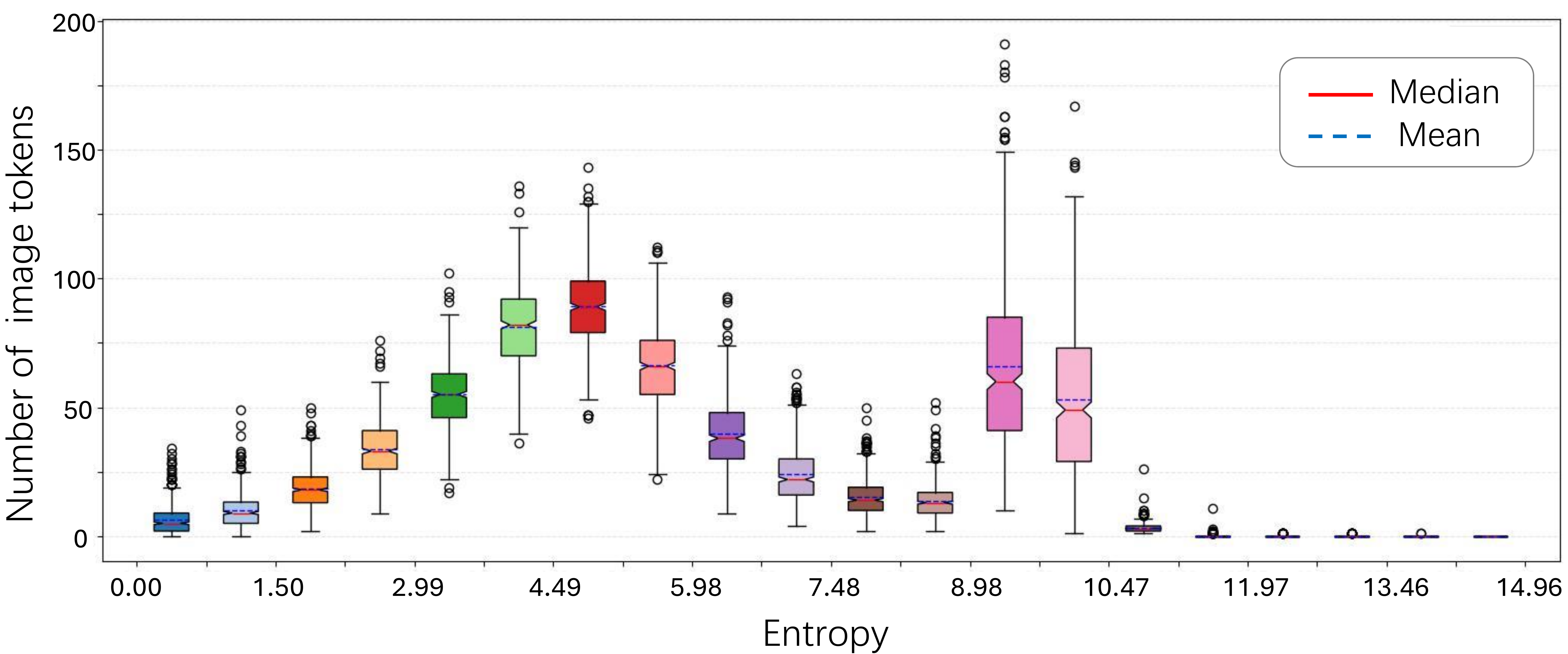}
      \label{subfig:entropy_count}
  }
    \caption{The next token predictions of image tokens. (a) depicts the input image: a cat behind a bottle in a room; (b) demonstrates the next token prediction of the image tokens accompanied by the entropy heatmap. The warmer color denotes lower entropy; (c) is the histogram of entropy $E(v_i)$ on the MSCOCO dataset. The entropy is divided into different bins and the number of visual tokens is counted for each bin. For each bin, the boxplot is utilized to show the statistics (\emph{e.g.,} mean, median, 25th and 75th percentile, \emph{etc.}) of token numbers across different images.}
    %is the grouped boxplot, showing the distribution of the number of image tokens across distinct entropy intervals counted on the MSCOCO~\cite{li2023evaluating} dataset. The rectangular box of each boxplot represents the interquartile range.}
  \label{fig:image_entropy}
  
\end{figure}

Let's take Figure \ref{fig:image_entropy} as an example. 
Figure \ref{subfig:cat_bottle} is the input image, which displays a cat and a bottle in a room, and 
Figure \ref{subfig:text_entropy} demonstrates the next token prediction of the image tokens, accompanied by the heat map plotted based on the entropy of predicted probabilities. The entropy calculation is formulated as 
\begin{equation}
    E(v_{i}) = -\sum_{t_v\in \Phi} p(t_{v}|v_{\leq i})\text{log}\space p(t_{v}|v_{\leq i}).
    \label{func_entropy}
\end{equation}
The warmer color indicates lower entropy. We can tell from the figure that the image tokens with low entropy are usually semantically meaningful. They make confident predictions consistent with the image content, like \textit{``le(leg)''}, \textit{``bott(bottle)''}, \textit{``cat''}, \textit{``flower''}, etc. While the image tokens with high entropy make meaningless predictions, such as \textit{``<0x0A>''}, \textit{``the''}, \textit{``.''}, and \textit{``</s>''}. 

Further, we plot the histogram of $E(v_i)$ on the MSCOCO dataset \cite{li2023evaluating}, as shown in Figure \ref{subfig:entropy_count}. We divide the entropy interval into different bins and count the number of visual tokens falling into each bin. As each bin may contain visual tokens from different images, we additionally show the statistics of token numbers across different images (\emph{e.g.,} the mean, median, 25th and 75th percentile, \emph{etc.}). 
Overall, the figure demonstrates an obvious bimodal distribution and the distribution generally holds for each image, indicating that quite a few image tokens make meaningless predictions with high entropy. 
To make text generation more dependent on informative visual information, we utilize $E(v_i)$ to measure the quality of the image tokens and remove the ambiguous or redundant ones (we denote the redundant visual token set as $\textbf{v}^b$) whose $E(v_i)$ is above a certain threshold $\tau$. 

%Further, we count the entropy distribution of image tokens across the POPE MSCOCO dataset \cite{li2023evaluating} to draw a more general conclusion. 
%Figure \ref{subfig:entropy_count} illustrates the distribution of the number of image tokens across distinct entropy intervals. 
%The x-axis divides entropy values into discrete bins, and the y-axis counts the number of image tokens. 
%For each bin, the boxplot shows the distribution of the number of visual tokens in its entropy interval. The rectangular box represents the interquartile range, containing the middle $50\%$ of the data. We can tell the consistency of the entropy distribution from it.
%Considering the distribution of the number of visual tokens across the entropy interval, the figure demonstrates an obvious bimodal distribution, indicating that quite a few image tokens make meaningless predictions with high entropy. 
%To make the text generation more dependent on informative visual information, we utilize $E(v_i)$ to measure the quality of the image tokens and remove the ambiguous or redundant ones (we denote the redundant visual token set as $\textbf{v}^b$) whose $E(v_i)$ is above a certain threshold $\tau$. 

The remaining image tokens are ``clear'' image tokens: $\textbf{v}^c = \{v^c_0, v^c_1, \dots,v^c_R\}$.  $\textbf{v}^c$ is a subset of origin image tokens set $\textbf{v}$ and each $v^c_i$ satisfies the entropy criterion, formulated as $E(v^c_i) < \tau$.

\subsection{Language Prior Rectification}

The language prior inherits bias from the Large Language Model backbone.
This bias will make the model lean toward generating unexpected words. 
To solve this problem, we propose a simple, yet effective rectification strategy based on Bayesian theory.

According to the Bayesian Formula, the posterior distribution is proportional to the prior distribution, and in the context of visual-relevant tasks, the ideal prior distribution is a uniform distribution over a certain vocabulary set. 
So, theoretically, we can rectify the defective prior by dividing it, and the result is proportional to the ideal posterior distribution. The rectification can be formulated as:
\begin{equation}
    p'(t_{n_t}) = \frac{1}{Z} \frac{p(t_{n_t}|\textbf{v}^{c}, \textbf{t})}{p(t_{n_t}|\textbf{t}) + \epsilon}, 
\end{equation}
where $p(t_{n_t}|\textbf{v}^{c}, \textbf{t})$ is the posterior distribution conditioned on the clear visual tokens $\textbf{v}^{c}$ and textual tokens $\textbf{t}$, $p(t_{n_t}| \textbf{t})$ is the prior distribution conditioned on the same textual tokens $\textbf{t}$. $p'(t_{n_t})$ is the rectified result and $Z$ is the normalization constant. We add a small positive number $\epsilon$ to the denominator to prevent computation overflow. 
In practice, to estimate the prior, we 
remove $\textbf{v}^c$ and utilize the text tokens and redundant visual tokens to estimate the probability of next-token vocabulary. We utilize the redundant visual tokens $\textbf{v}^b$ as they do not introduce any informative visual information into the prior, but can help us achieve a sample-specific rectification.

Given that the ideal prior is a uniform distribution over a vocabulary subset, and dividing by a fairly small prior probability may lead to unreasonable predictions, it is not appropriate to rectify the distribution over the whole vocabulary set. Following \cite{li2022contrastive,chuang2023dola,huo2024self,leng2024mitigating}, we employ an adaptive plausibility constraint to calibrate the entire output distribution, preventing implausible outputs from the rectified distribution. Specifically, we employ the following strategy to rectify the prior, \emph{i.e.,}
\begin{equation}
p'(t_{n_t}) = 
\begin{cases}
p'(t_{n_t}) & \text{if}  \quad t_{n_t} \in \Phi'(t_{n_t}) \\
0     &  \text{if} \quad t_{n_t} \notin \Phi'(t_{n_t})
\end{cases} 
\end{equation}
\begin{equation}
%\left.
%\begin{array}{cc} 
    \Phi'(t_{n_t}) = \left\{ t_{n_t} | t_{n_t} \in \Phi, p(t_{n_t}|\textbf{v}^{c}, \textbf{t}) > \mu \cdot \underset{\omega}{\text{max}} p(\omega|\textbf{v}^{c}, \textbf{t})\right\}, \\ \\
    %p'(y_i) = 0, \quad \text{if} \; \;  y_i \notin \Phi'(y) 
%\end{array}
%\right.
\end{equation}
where $\Phi'$ is a subset of the vocabulary $\Phi$, containing the words with high posterior probability values. $\mu$ controls the strength of vocabulary truncation, and larger $\mu$ indicating more aggressive truncation that retains only high-probability words.

It is worth noting that our method is different from the contrastive decoding methods~\cite{leng2024mitigating,wang2024mitigating,huo2024self}. 
Previous contrastive decoding methods rectify the distribution by subtracting the logits of distorted predictions from the original logits.
In contrast, our method operates on the predicted probability (\emph{i.e.,} dividing the prior by posterior), which is more aligned with the Bayesian theory and yields better performance (see Section~\ref{subsec:ablation} for comparison). 

\begin{figure}[h]
    \raggedright
  \subfloat[]{
      \includegraphics[width=0.23\textwidth]{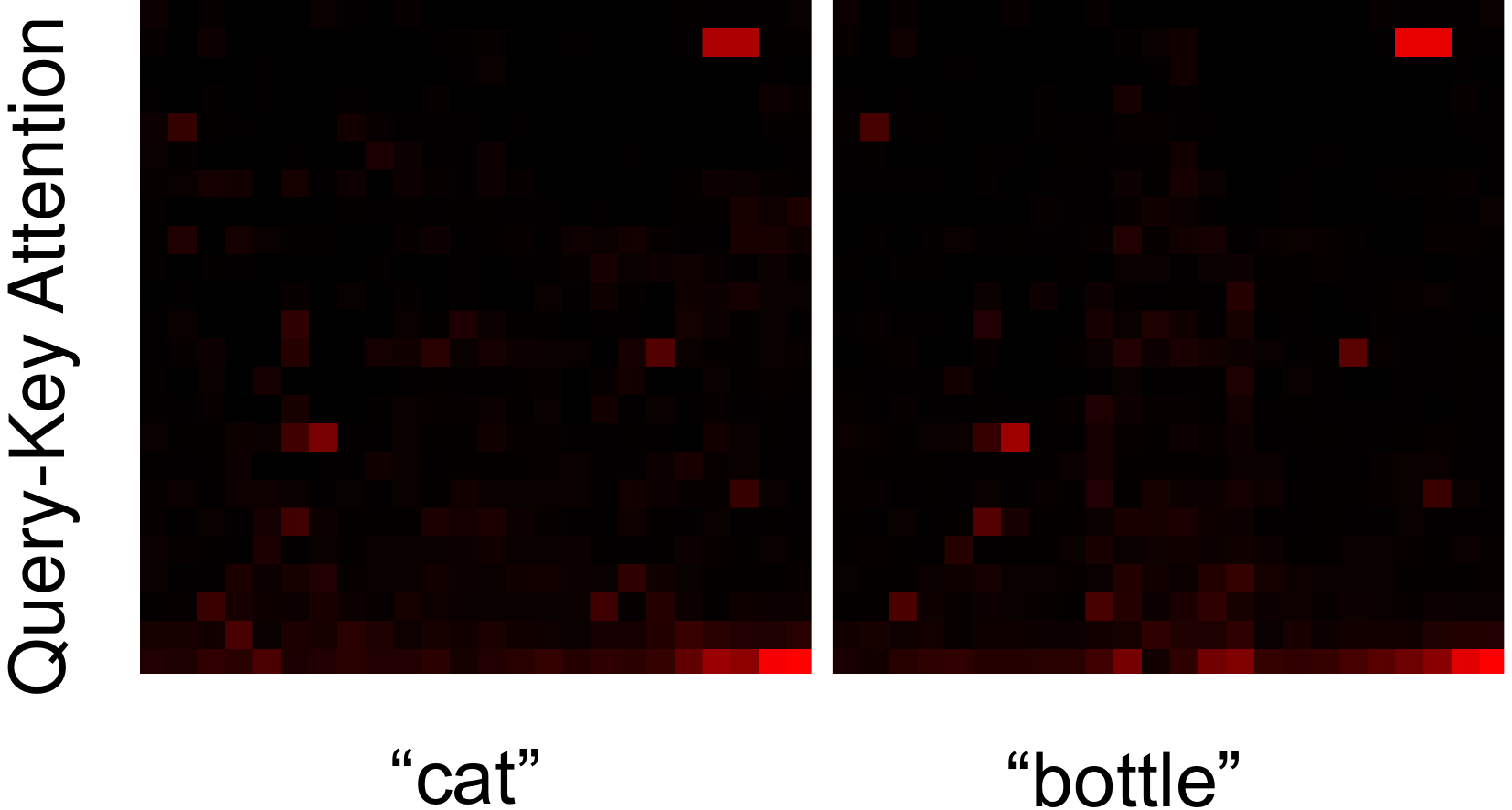}
      \label{subfig:qk_attn}
  }
  \subfloat[]{
      \includegraphics[width=0.23\textwidth]{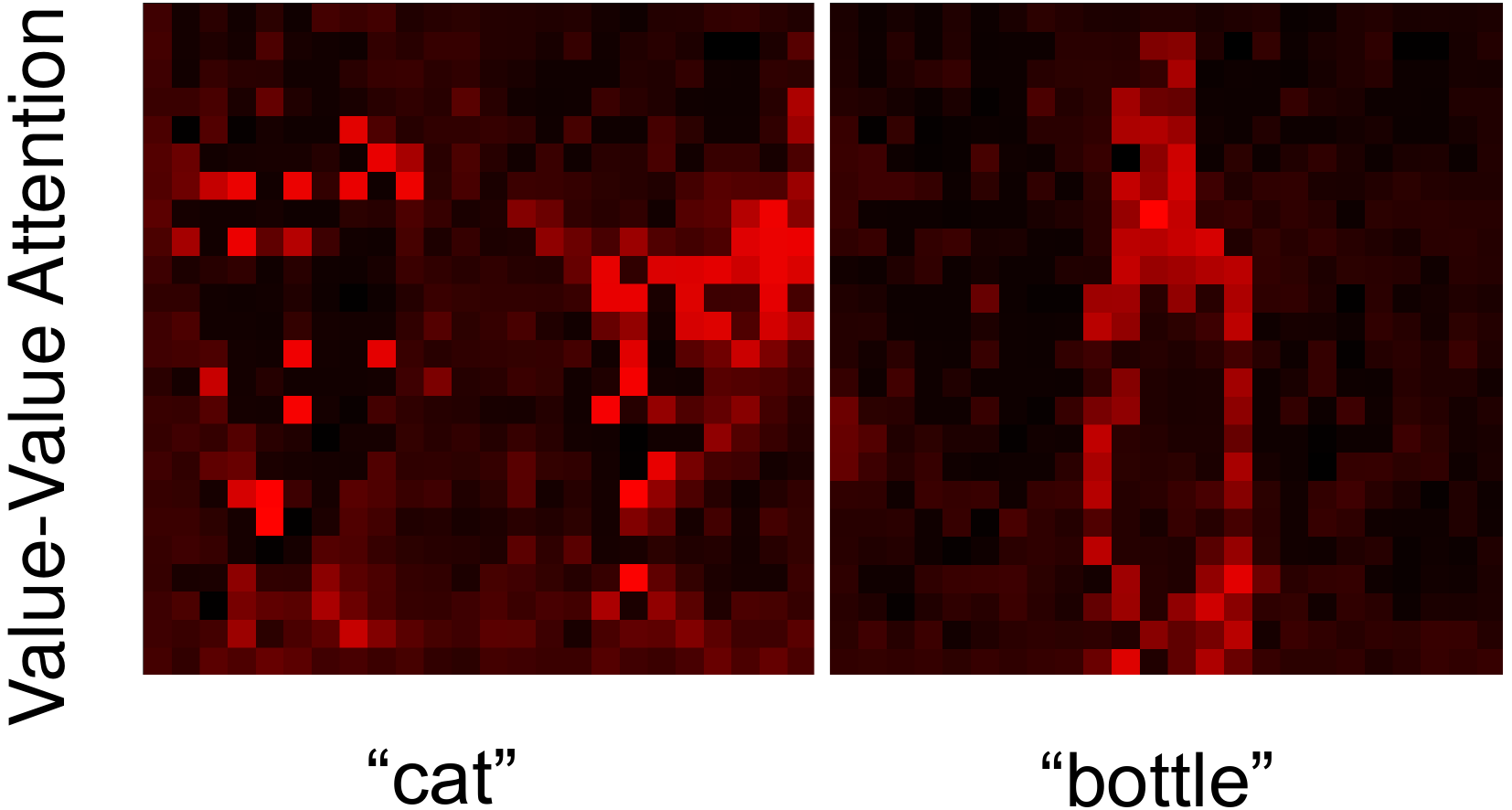}
      \label{subfig:vv_attn}
  }
    \caption{The text-to-image query-key attention maps (a) and value-value attention maps (b).}
    \label{fig:attn}
\end{figure}

\begin{figure}[h]
      \includegraphics[width=0.48\textwidth]{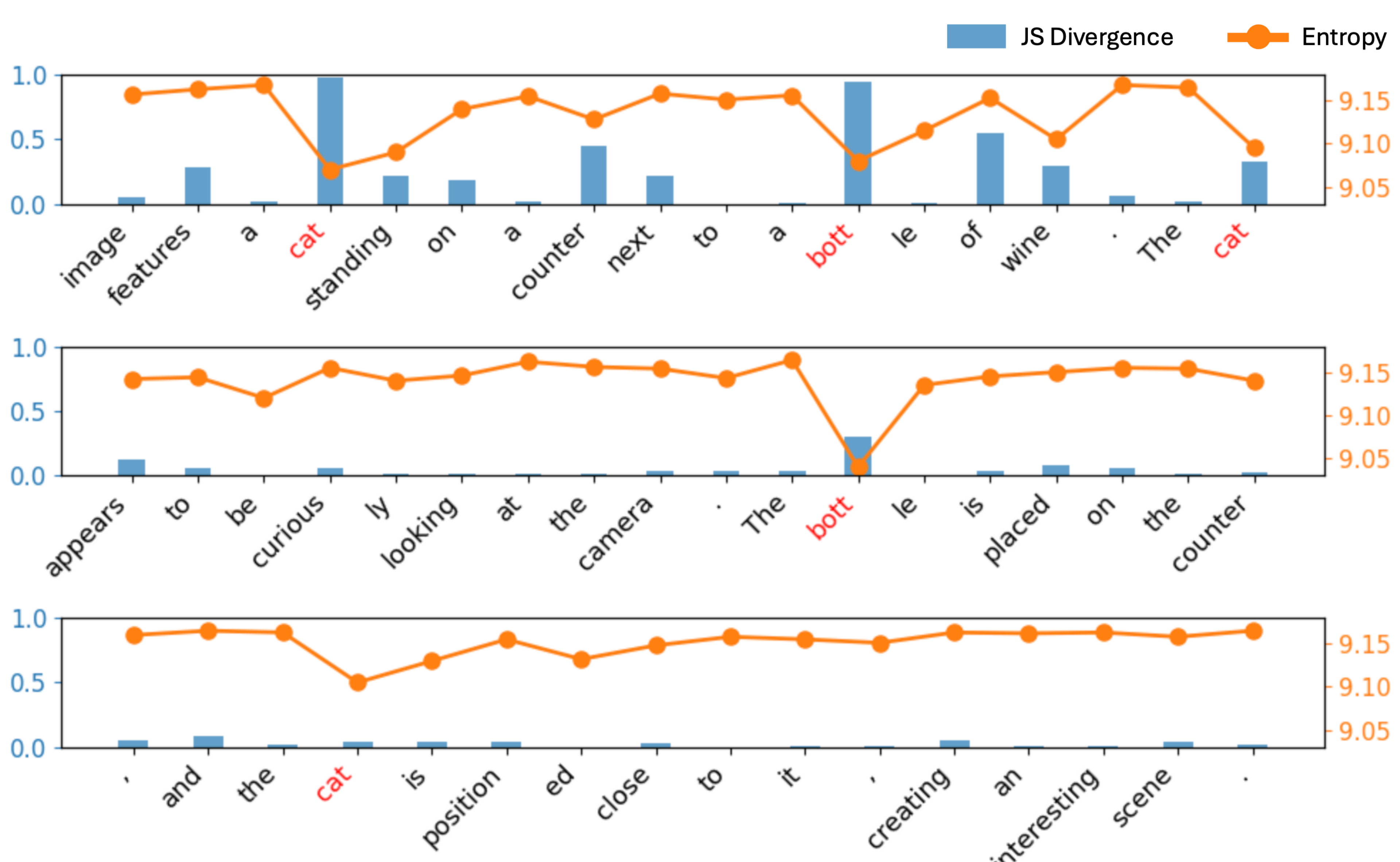}
    \caption{Generated caption for Figure \ref{subfig:cat_bottle}. The entropy of the text-to-image value-value attention map and JS divergence between the posterior and prior distribution.} 
  \label{fig:js_div}
\end{figure}

%\subsection{Early Stopping based on Posterior Collapse Discovery}
\subsection{Collapse-Aware Early Stopping}
\label{early stopping}
When the posterior collapses into the prior, the inference process is only conditioned on textual information and incurs hallucinations. 
So, it is necessary for us to detect the posterior collapse, making an early stopping before the posterior collapses into the prior. 

When generating visual-relevant text tokens, the posterior and prior distributions should be different, because the former has access to visual information, while the latter can only make a guess based on the text context. 
However, when the posterior collapse occurs, the posterior inference also becomes conditioned solely on the textual information, leading it to make similar predictions as the prior for the visual-relevant text tokens. 
To quantify the degree of posterior collapse, we can measure the similarity of the posterior and prior distribution of these visual-relevant text tokens.

First, we should locate the visual-relevant text tokens. Recently, some works \cite{bousselham2024grounding,li2023clip,wang2024sclip} on CLIP \cite{radford2021learning} apply self-self attention to emphasize semantic correlations. Inspired by these works, we attempt to utilize self-self attention mechanisms in LVLM to quantify the correlation between text and visual tokens. 
Through an analysis of value-value attention across text and image tokens, we discover that there exists significant semantic consistency between the value features of text tokens and the image tokens. The value-value attention across text and image tokens can be formulated as
\begin{equation}
\label{func:vv}
    \alpha(t_i, \textbf{v}) =\text{softmax}\left( \frac{(h_{n_s+i}W_V)(h_{<n_s}W_V)^\top}{\sqrt{d}} \right),
\end{equation}
where $W_V$ is value projection matrix and $d$ is channel dimension.

As shown in Figure \ref{fig:attn}, we perform query-key and value-value attention using the text tokens \textit{``cat''} and \textit{``bottle''} with the image tokens from Figure \ref{subfig:cat_bottle}, respectively.
Figures \ref{subfig:qk_attn} and \ref{subfig:vv_attn} visualize the corresponding query-key and value-value attention maps, averaged across all layers and heads. The red color stands for a high attention score. 
In contrast to query-key attention, value-value attention maps highlight the outline of the corresponding object, demonstrating a strong alignment between the value features of text tokens and their semantically related image tokens.
Then the entropy of the value-value attention map can reflect the degree of correlation between visual tokens and text tokens.  We then leverage the entropy to select the visual-relevant textual tokens.

In practice, the $i$-th generated token will serve as the $(i+1)$-th input text token. 
We compute the entropy of its value-value attention map with the clear image tokens, denoted as $E(\alpha(t_{i+1},\textbf{v}^c))$. We then calculate the change in entropy relative to the previous token, \emph{i.e.,} $\Delta E = E\left(\alpha(t_{i+1},\textbf{v}^c)) - E(\alpha(t_{i},\textbf{v}^c)\right) $. A significant negative change (defined as $ \Delta E < -\delta$, where $\delta$ is a predefined positive threshold) means the entropy significantly decreases, and we treat $t_{i+1}$ as visual-relevant token.

After identifying the visual-relevant tokens, we compute the similarity between the posterior and prior distributions over these tokens to quantify the degree of posterior collapse. We leverage JS divergence to measure the distance, \emph{i.e.,}
\begin{equation}
%\begin{array}{cc}
    \text{JS}(A||B)=\frac{1}{2}\text{KL}(A||B)+\frac{1}{2}\text{KL}(B||A),
    %\text{KL}(A||B)=\sum_{i}A(i)\log\left(\frac{A(i)}{B(i)}\right)
%\end{array}
\end{equation}
where $KL(\cdot||\cdot)$ denotes KL Divergence, $A$ stands for posterior distribution  $p(t_{n_t}|\textbf{v}^{c}, \textbf{t})$ and $B$ stands for prior distribution $p(t_{n_t}| \textbf{t})$.  

Typically, before the hallucinations occur, the model tends to generate detailed descriptions by repeatedly referring to visual objects present in the image. 
As demonstrated by Figure \ref{fig:js_div}, the visual-relevant tokens, \textit{"cat"} and \textit{"bott"} recur in the generated sequence, and their JS divergences decrease with progressive sequence extension.
So we can track the JS divergence shifts of recurring visual-relevant tokens and apply early stopping based on them.

In practice, we maintain a dynamic dictionary $D$ where the word of the visual-relevant token is the key and the corresponding JS distance is the value. 
Specifically, for each visual-relevant token $t_i$, we store its initial JS divergence $JS_f$ at its first occurrence and the latest JS divergence $JS_l$ at the last occurrence. 

Then, we compute the divergence shift for each recurring visual-relevant token, \emph{i.e.,}  $\Delta JS = \max(JS_f - JS_l, 0)$, and use the mean value of $\Delta JS$ over all the visual-relevant tokens in $D$ (denoted as $\overline{\Delta JS}$) as a dynamic scaling factor for the <EOS> logit, \emph{i.e.,} 
\begin{equation}
   \text{logit}_{\text{eos}} = \text{logit}_{\text{eos}} \cdot \left(1 + \lambda \cdot \overline{\Delta JS}\right), 
\end{equation} 
where $\lambda$ is a hyperparameter that controls the scaling magnitude. 
As the text generation proceeds, the $\text{logit}_{\text{eos}}$ is adaptively adjusted based on the decline in JS divergence, allowing the model to terminate the generation before the posterior collapses into the prior.
Further, to ensure grammatical accuracy, we limit logit scaling operations to tokens that immediately follow the punctuation marks. 

In this way, we adjust the termination probability based on the decrease in JS divergence of visual-relevant candidates, effectively mitigating the hallucinations caused by posterior collapse.

\subsection{Decoding pipeline}
The proposed three strategies: redundant visual token pruning, language prior rectification, and collapse-aware early stopping, can be implemented in one forward pass, which doesn't introduce much additional inference cost compared to vanilla LVLM inference.

Vanilla LVLM inference can be divided into two stages: prefill and decode. In the prefill stage, the system, visual, and instruction tokens are processed, and their KV-cache is stored for decoding. In the decode stage, new text tokens will be auto-regressively generated with the help of the KV-cache.
Back to our method, in the prefill stage, we leverage the last hidden states of visual tokens to calculate their entropy, according to which we separate redundant visual tokens from clear tokens in the KV-cache and perform token pruning. 
In the decode stage, benefiting from the separated KV-cache, we can perform two parallel decodings to calculate the posterior and prior, respectively. Based on those two outputs, we perform language prior rectification and collapse-aware early stopping.

Thus, our method does not introduce too much computation cost and is still computation-efficient (see Section \ref{efficiency} for  comparisons). %Quantitative experimental results supporting this claim can be found in Section \ref{efficiency}.

%So our method is more computation-efficient than other hallucination mitigating methods (see Section \ref{efficiency} for detailed comparisons). %Quantitative experimental results supporting this claim can be found in Section \ref{efficiency}.

\section{Experiments}
\label{sec:experiment}
\subsection{Benchmarks}
We conduct experiments on both discriminative and generative tasks for hallucination evaluation, and a comprehensive benchmark MME is selected to test the general ability. 

\textbf{POPE.} POPE \cite{li2023evaluating} (Polling-based Object Probing Evaluation) is a benchmark designed to evaluate the ability of LVLMs to accurately determine the presence or absence of specific objects in images. 
The referred objects are selected by 
three different sampling options: random, popular, and adversarial to cover different difficulty levels.
To quantify performance, POPE employs standard binary classification metrics: accuracy, precision, recall, and F1-score, aligning with established evaluation practices in the field.

\textbf{CHAIR.} CHAIR \cite{rohrbach2018object} (Caption Hallucination Assessment with Image Relevance) evaluates object hallucinations in image captions by comparing extracted objects with ground-truth labels. It measures accuracy at instance ($\text{CHAIR}_\text{I}$) and sentence ($\text{CHAIR}_\text{S}$) levels. 
$\text{CHAIR}_\text{I}$ is the ratio of the number of hallucinated objects to the total number of objects mentioned across all captions and $\text{CHAIR}_\text{S}$ is the ratio of the number of captions containing at least one hallucinated object to the total number of captions. Recall is used to measure the completeness of the captions. Following previous works \cite{huang2024opera}, we randomly select 500 images from the MSCOCO dataset \cite{liu2023mitigating}, and use the prompt: ``Please help me describe the image in detail.''

\textbf{MME.} MME \cite{fu2306mme} (Multimodal Large Language Model Evaluation) quantitatively assesses the perceptual and cognitive abilities of LVLMs across fourteen subtasks, including existence, count, position, color, posters, celebrity, scene, landmark, artwork, and OCR for perception and commonsense reasoning, numerical calculation, text translation, and code reasoning for cognition.

\begin{table*}[t]
\centering
\setlength{\tabcolsep}{8pt} % 减少列间距
\begin{tabular}{l*{12}{c}}
\toprule
\multirow{2}{*}{Method} & \multicolumn{4}{c}{MSCOCO} & \multicolumn{4}{c}{GQA} & \multicolumn{4}{c}{A-OKVQA} \\
\cmidrule(lr){2-5} \cmidrule(lr){6-9} \cmidrule(lr){10-13}
 & Acc & Prec & Rec & F1  & Acc & Prec & Rec & F1  & Acc & Prec & Rec & F1  \\
\midrule

Vanilla \cite{liu2024improved} & 85.5 & 88.3 & 82.4 & 85.1 & 76.3 & 70.5 & 93.5 & 80.1 & 78.2 & 73.1 & 91.5 & 81.1 \\
OPERA \cite{huang2024opera} & 84.8 & 82.6 & \textbf{88.8} & \textbf{85.5} & 76.7 & 70.0 & 96.1 & 80.8 & 78.3 & 72.0 & 94.9 & 81.6 \\
ICD \cite{wang2024mitigating} & 84.2 & 82.2 & 88.0 & 84.9 & 76.9 & 70.0 & 96.4 & 80.9 & 78.4 & 72.0 & 94.9 & 81.7 \\
VCD \cite{leng2024mitigating} & 83.6 & 81.1 & 88.6 & 84.5 & 75.6 & 68.8 & 96.1 & 80.0 & 77.4 & 71.1 & 94.6 & 81.0 \\
SID \cite{huo2024self} & 85.3 & 84.6 & 86.9 & 85.6 & 79.3 & 73.1 & 94.7 & 82.3 & 80.4 & 74.7 & 93.6 & 82.9 \\
DeCo \cite{wang2024mllm} & 84.0 & 79.1 & 93.1 & 85.4 & 74.0 & 66.8 & \textbf{98.8} & 79.5 & 76.4 & 68.9 & \textbf{98.5} & 80.9 \\
CausalMM \cite{zhou2024mitigating} & 85.6 & 88.3 & 82.4 & 85.2 & 84.1 & 83.5 & 85.5 & 84.4 & 84.5 & 83.7 & 86.3 & 84.9 \\
Ours & \textbf{85.6} & \textbf{88.3} & 82.4 & 85.2 & \textbf{84.2} & \textbf{83.6} & 85.6 & \textbf{84.5} & \textbf{84.5} & \textbf{83.7} & 86.4 & \textbf{84.9} \\
\bottomrule
\end{tabular}
\caption{Comparison on discriminative task: POPE evaluation on MSCOCO, GQA, and A-OKVQA dataset.
The scores are averaged under the three sub-tasks: random, popular, and adversarial. The best results are in bold. Our method consistently presents a high performance.}
\label{pope}
\end{table*}

\subsection{Baselines}
We compare our methods with previous training-tree methods for mitigating hallucinations, as outlined below:
\textbf{OPERA} \cite{huang2024opera} dynamically penalizes overconfident tokens during beam search and incorporates a rollback strategy for token selection to mitigate hallucination;
\textbf{ICD} \cite{wang2024mitigating}  mitigates object hallucinations in LVLMs by contrasting output distributions from the original and negative prompted inference;
\textbf{VCD} \cite{leng2024mitigating} reduces the hallucination by enhancing visual information through contrasting output distributions from original and distorted visual inputs;
\textbf{SID} \cite{huo2024self} adaptively mitigates vision and text association hallucinations during auto-regressive decoding by contrastive decoding with the least important vision tokens;
\textbf{DeCo} \cite{wang2024mllm} mitigates hallucination by adaptively selecting the appropriate preceding layers and proportionally integrating knowledge into the final layer to adjust the output logits;
\textbf{CausualMM} \cite{zhou2024mitigating} applies structural causal modeling to MLLMs, treating modality priors as a confounder between attention mechanisms and output.

\subsection{Implementation Detail}
The entropy threshold for redundant visual token pruning $\tau$ is set to 7.48. %, which is the valley of the bimodal distribution.
The threshold for visual-relevant tokens selection $\delta$ is set to 0.05, and the scaling ratio for early stopping $\lambda$ is set to 1.5.
To adaptively handle different kinds of tasks, the probability truncation ratio $\mu$ for language prior rectification is set to 0.1 for POPE \cite{li2023evaluating}, 0.9 for CHAIR \cite{rohrbach2018object}, and 0.85 for MME  \cite{fu2306mme}. 
We conduct experiments based on two LVLMs: LLaVA-1.5 \cite{liu2024improved} and Qwen-2.5-vl \cite{bai2025qwen2}. All the experiments are conducted on A100 GPUs.

% \subsection{Experimental Results }
\subsection{Comparison with previous state-of-the-arts}
For a fair comparison, except OPERA \cite{huang2024opera} (designed for beam search), all methods employ greedy decoding for sampling. ``Vanilla'' stands for the original model.
For all baselines, we use their default hyperparameters. The experiments are conducted on LLaVA-1.5-7b unless otherwise stated.

\textbf{Discriminative task.}
We evaluate the POPE \cite{li2023evaluating} task performance on three datasets, including MSCOCO \cite{lin2014microsoft}, A-OKVQA \cite{schwenk2022okvqa}, and GQA \cite{hudson2019gqa}. Table \ref{pope} reports the average scores under the three settings, e.g., random, popular, and adversarial. EVRB demonstrates a state-of-the-art performance in mitigating object hallucinations on discriminative tasks. Note that though CausalMM performs comparably, it is a discriminative-task-specific method, performing poorly in generation and comprehensive tasks.

\setlength{\tabcolsep}{3.5pt}
\begin{table}[h]
\centering
\begin{tabular}{lcccc}
\toprule
Method & $\text{CHAIR}_\text{S}\downarrow$ & $\text{CHAIR}_\text{I}\downarrow$ & Recall  & Length  \\
\midrule
Vanilla \cite{liu2024improved} & 50.0 & 14.1 & \textbf{80.4} & 98.2  \\

OPERA \cite{huang2024opera} & 50.6 & 15 & 79 & 95.6  \\

ICD \cite{wang2024mitigating} & 49.6 & 14.0 & 80.2 &  101.4 \\

VCD \cite{leng2024mitigating} & 52.4 & 14.4 & 79.7 & 98.9  \\

SID \cite{huo2024self} & 49.8 & 13.6 & 78.9 & 95.0  \\

DeCo \cite{wang2024mllm} & 41.4 & \textbf{11.0} & 73.4 & 100.1  \\

CausalMM \cite{zhou2024mitigating} & 50.4 & 14.5 & 78.8 & 99.7  \\

Ours & \textbf{39.8} & 12.8 & 78.5 & 87.7 \\
\bottomrule
\end{tabular}
\caption{CHAIR hallucination evaluation results. % based on LLaVA-1.5. 
Smaller CHAIR corresponds to less hallucination, and higher recall means better completeness. The best results are in bold. Our method has a low hallucination rate with comparable recall.}
\label{chair}
\end{table}

\textbf{Generative task.} Besides the evaluation of the discriminative task, we test the methods on the generative task using the CHAIR \cite{rohrbach2018object}. The max new tokens hyperparameter is set to 512. Table \ref{chair} demonstrates that our model performs favorably against the previous state-of-the-art, mitigating hallucination while preserving the grounding ability to a great extent. Though DeCo \cite{wang2024mllm} performs better on the $\text{CHAIR}_\text{I}$, it demonstrates a low recall score. Notably, the $\text{CHAIR}_\text{I}$ score is negatively correlated with the total number of objects mentioned in captions, meaning reduced recall can paradoxically improve this metric. This suggests that DeCo \cite{wang2024mllm} prioritizes reducing hallucinations at the expense of captioning capabilities.

\begin{figure}[h]
  \centering
  \includegraphics[width=\linewidth]{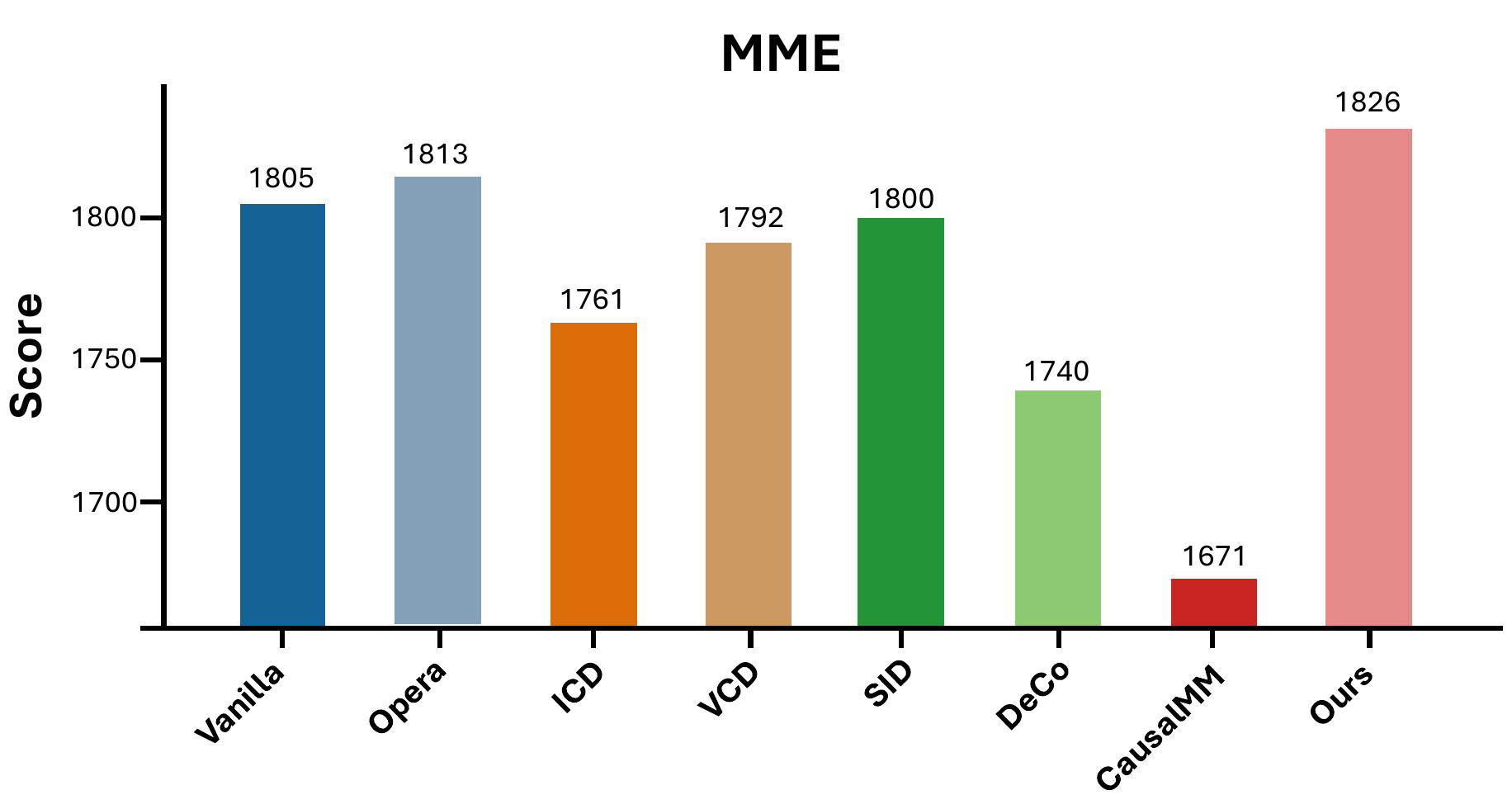} 
  \caption{Comparison on comprehensive task: MME evaluation. Our method generally improves the performance.} 
  \label{mme}
\end{figure}

\textbf{Comprehensive task.} We leverage MME  \cite{fu2306mme} to test the comprehensive ability. The comparisons with existing methods are shown in Figure \ref{mme}. Although can alleviate hallucinations to a certain extent, most previous approaches undermine the comprehensive performance of the LVLMs. In contrast, our method can generally improve the performance of the model.

% \begin{table}[h]
% \centering
% \label{mme}
% \begin{tabular}{lccc}
% \toprule
% \textbf{Method} & \textbf{Perception} & \textbf{Cognition} & \textbf{Total Score} \\
% \midrule
% Vanilla & 1482.865 & 322.5 & 1805.365 \\
% opera      & 1479.497 & 333.571 & 1813.068 \\
% Sid        & 1481.594 & 318.214 & 1800.808 \\
% Vcd        & 1469.271 & 322.857 & 1792.128 \\
% Icd        & 1398.101 & 290.357 & 1688.458 \\
% Deco       & 1461.779 & 277.5 & 1739.279 \\
% CausalMM   & 1418.824 & 251.786 & 1670.610 \\
% Ours      & 1453.95 & 366.43 & \textbf{1820.38} \\
% \bottomrule
% \end{tabular}
% \caption{}
% \end{table}

\subsection{Ablations}
\label{subsec:ablation}
\textbf{Effect of each component.}
We conduct ablation studies to show the contribution of redundant visual token pruning, language prior rectification, and collapse-aware early stopping to the performance. Table \ref{ablation:components_chair} shows that all of the designs can mitigate the hallucination on CHAIR \cite{rohrbach2018object} benchmark. %and the collapse-aware early stopping strategy makes the main contribution. 
Further, compared to only applying the early stopping strategy, the complete method has a higher recall score. This demonstrates that the token pruning combined with prior rectification can benefit the LVLM's grounding ability. For POPE \cite{li2023evaluating}  benchmark, given that the collapse-aware early stopping strategy is irrelevant to discriminative tasks, we conduct ablation for redundant visual token pruning and language prior rectification. Table \ref{ablation:components_pope} shows that both strategies contribute to hallucination reduction. It is worth noting that the prior rectification strategy makes the main contribution when dealing with the hard discriminative tasks (POPE on A-OKVQA and GQA datasets).

% We do ablation studies to show the contribution of each component to the performance.
% We do ablation on redundant visual token pruning, prior rectification, and posterior-aware early stopping. Table \ref{ablation:components_chair} shows that all of the designs can mitigate the hallucination in generative tasks and the posterior-aware early stopping strategy makes the main contribution. Further, compared to only applying the early stop strategy, the complete method has a higher recall score. This demonstrates that the token pruning combined with prior rectification can benefit the LVLM's grounding ability. For POPE \cite{li2023evaluating}  benchmark, given that the posterior-aware early stopping strategy has no concern with discriminative tasks, we do ablation for redundant visual token pruning and prior rectification. Table \ref{ablation:components_pope} shows both strategies contribute to hallucination reduction. It is worth noting that, the prior rectification strategy makes the main contribution when dealing with the hard discriminative tasks (POPE on A-OKVQA and GQA datasets).

\setlength{\tabcolsep}{0.7pt}
\begin{table}[h]
\centering
\begin{tabular}{cccccc}
\toprule
\scalebox{0.8}{$\begin{array}{c} \text{Token} \\ \text{Pruning} \end{array}$} 
& \scalebox{0.8}{$\begin{array}{c} \text{Prior} \\ \text{Rectification} \end{array}$} 
& \scalebox{0.8}{$\begin{array}{c} \text{Early} \\ \text{Stopping} \end{array}$} &  
$\text{CHAIR}_\text{S}\downarrow$ & $\text{CHAIR}_\text{I}\downarrow$ & Recall  \\
\midrule
\checkmark &  &  & 48.6 & 14.0 & 79.7  \\
& \checkmark &   & 49.0 & 14.0 & \textbf{79.9} \\
&  & \checkmark & 43.4 & 13.4 & 77.4  \\
\checkmark& \checkmark & \checkmark & \textbf{39.8} & \textbf{12.8} & 78.5  \\
  
\bottomrule
\end{tabular}
\caption{Ablation on the components of our design under CHAIR benchmark. Each component contributes to the superior hallucination mitigation effect.}
\label{ablation:components_chair}
\end{table}

\setlength{\tabcolsep}{2.2pt}
\begin{table}[h]
\centering
\begin{tabular}{cccccccc}
\toprule
\multirow{2}{*}{\scalebox{0.8}{$\begin{array}{c} \text{Token} \\ \text{Pruning} \end{array}$}} & \multirow{2}{*}{\scalebox{0.8}{$\begin{array}{c} \text{Prior} \\ \text{Rectification} \end{array}$}} & \multicolumn{2}{c}{MSCOCO} & \multicolumn{2}{c}{GQA} & \multicolumn{2}{c}{A - OKVQA} \\
\cmidrule(lr){3-4} \cmidrule(lr){5-6} \cmidrule(lr){7-8}
 & & acc & F1 & acc & F1 & acc & F1 \\
\midrule
\checkmark & & 85.0 & \textbf{85.6}	&77.4&81.1&78.9&82.0 \\
& \checkmark & 84.8 &	84.8&83.0	&84.0&83.1 & 84.2 \\
\checkmark & \checkmark & \textbf{85.6} & 85.2 & \textbf{84.2} & \textbf{84.5} & \textbf{84.5} & \textbf{84.9} \\
\bottomrule
\end{tabular}
\caption{Ablation on the components of our design under POPE benchmark. Each component contributes to the superior hallucination mitigation effect.}
\label{ablation:components_pope}
\end{table}

% \begin{figure*}[t]
%   \centering
%   \includegraphics[width=\linewidth]{graphs/case_study.pdf} 
%   \caption{A case study on image caption. Both DeCo and our method can eliminate the hallucination, while Deco fails to generate captions for all the ground-truth objects. The word in blue means the ground-truth object, while the word in red means the hallucination.} 
%   \label{case_study}
% \end{figure*}

\setlength{\tabcolsep}{3pt}
\begin{table*}[t]
\centering
\caption{POPE, CHAIR, and MME evaluation results based on Qwen2.5-vl-7b and LLaVA-1.5-13b.}
\begin{tabular}{clccccccccc}
\toprule
\multicolumn{2}{c}{Setting} & \multicolumn{2}{c}{POPE(COCO)} & \multicolumn{2}{c}{POPE(GQA)} & \multicolumn{2}{c}{POPE(A-OKVPA)} & \multicolumn{2}{c}{CHAIR} & {MME} \\
\cmidrule{1-2} \cmidrule(lr){3-4} \cmidrule(lr){5-6} \cmidrule(lr){7-8} \cmidrule(lr){9-10} 
 {Model} & {Decoding} & {Acc} & {F1} & {Acc} & {F1} & {Acc} & {F1} & {$\text{CHAIR}_\text{S}\downarrow$} & {$\text{CHAIR}_\text{I}\downarrow$} & {Score$\uparrow$}\\
\midrule

\multirow{4}{*}{\rotatebox[origin=c]{0}{Qwen2.5-vl-7b \cite{bai2025qwen2}}} & Vanilla & 0.838 & 0.809 & 0.833 & 0.828 & 0.846 & 0.838 & 40.4 & 10.0 & 2317 \\
& ICD \cite{wang2024mitigating}     & 0.842 & 0.816 & 0.840 & 0.834 & 0.853 & 0.847 & 41.0 & 12.3 & 2259 \\
& VCD \cite{leng2024mitigating}     & 0.864 & 0.847 & 0.842 & 0.842 & 0.861 & 0.859 & 42.2 & 11.8 & 2335 \\
& EVRB    & \textbf{0.880} & \textbf{0.869} & \textbf{0.855} & \textbf{0.859} & \textbf{0.869} & \textbf{0.868} & \textbf{34.0} & \textbf{10.0} & \textbf{2345} \\
\midrule

\multirow{4}{*}{\rotatebox[origin=c]{0}{LLaVA1.5-13b \cite{liu2024improved}}}&Vanilla & 0.854 & \textbf{0.864} & 0.757 & 0.802 & 0.772 & 0.811 & 46.2 & 12.7 & 1792 \\
&ICD \cite{wang2024mitigating}     & 0.844 & 0.857 & 0.749 & 0.797 & 0.757 & 0.801 & 46.8 & 12.6 & \textbf{1826} \\
&VCD \cite{leng2024mitigating}     & 0.845 & 0.856 & 0.757 & 0.802 & 0.770 & 0.808 & 51.6 & 14.4 & 1808 \\
&EVRB    & \textbf{0.860} & 0.852 & \textbf{0.857} & \textbf{0.859} & \textbf{0.846} & \textbf{0.848} & \textbf{41.8} & \textbf{11.9} & 1814 \\

\bottomrule
\end{tabular}
\label{more_model}
\end{table*}

\textbf{Effect of using $\textbf{v}^b$ in prior estimation.} Ambiguous or redundant visual tokens $\textbf{v}^b$ are utilized in prior estimation to enable more sample-specific rectifications. Here we verify its effectiveness. As shown in Table \ref{ablation:amb_visual}, without $\textbf{v}^b$, though there is a slight improvement on the MSCOCO dataset, the performance on A-OKVQA and GQA datasets deteriorates significantly, showing the necessity to introduce $\textbf{v}^b$ when dealing with hard discriminative tasks.

\textbf{Bayesian Prior Rectification \emph{vs.} Logit Adjustment.} Another key design for prior rectification is the Bayesian-based rectification. We verify its effectiveness by comparing it with the conventional logit adjustment strategy adopted in SID \cite{huo2024self}. As shown in Table \ref{ablation:bayes_rec}, the performance of our design exceeds the logit adjustment strategy over all three datasets.

% \setlength{\tabcolsep}{3pt}
% \begin{table}[h]
% \centering
% \caption{Ablation study on ambiguous visual information exploitation and Bayesian-based rectification.}
% \begin{tabular}{cccccccc}
% \toprule
% \multirow{2}{*}{\scalebox{0.8}{$\begin{array}{c} \text{Bayesian} \\ \text{Rectification} \end{array}$}} & \multirow{2}{*}{\scalebox{0.8}{$\begin{array}{c} \text{Ambiguous} \\ \text{Visual Info} \end{array}$}} & \multicolumn{2}{c}{MSCOCO} & \multicolumn{2}{c}{GQA} & \multicolumn{2}{c}{A - OKVQA} \\
% \cmidrule(lr){3-4} \cmidrule(lr){5-6} \cmidrule(lr){7-8}
%  & & acc & F1 & acc & F1 & acc & F1 \\
% \midrule
% \checkmark & &  \textbf{84.9} & \textbf{85.3} & 79.4 & 82.4 & 80.3 & 82.9  \\
%  & \checkmark & 84.8 & 84.7 & 79.6 & 80.3 & 81.0 & 83.5 \\
% \checkmark & \checkmark & 84.8 & 84.8 & \textbf{83.0}	& \textbf{84.0} & \textbf{83.1} & \textbf{84.2} \\
% \bottomrule
% \end{tabular}
% \label{ablation:piror_rectification}
% \end{table}

\setlength{\tabcolsep}{6pt}
\begin{table}[h]
\centering
\caption{Effect of using $\textbf{v}_b$ in prior estimation. Only language prior rectification module is utilized.}
\begin{tabular}{ccccccc}
\toprule
\multirow{2}{*}{\scalebox{0.9}{$\begin{array}{c} \text{Ambiguous} \\ \text{Visual Info} \end{array}$}} & \multicolumn{2}{c}{MSCOCO} & \multicolumn{2}{c}{GQA} & \multicolumn{2}{c}{A - OKVQA} \\
\cmidrule(lr){2-3} \cmidrule(lr){4-5} \cmidrule(lr){6-7}
 & acc & F1 & acc & F1 & acc & F1 \\
\midrule
w. &  \textbf{84.9} & \textbf{85.3} & 79.4 & 82.4 & 80.3 & 82.9  \\

w/o. & 84.8 & 84.8 & \textbf{83.0}	& \textbf{84.0} & \textbf{83.1} & \textbf{84.2} \\
\bottomrule
\end{tabular}
\label{ablation:amb_visual}
\end{table}

\setlength{\tabcolsep}{3pt}
\begin{table}[h]
\centering
\caption{Effect of Bayesian-based rectification. Only the language prior rectification module is utilized.}
\begin{tabular}{lcccccc}
\toprule
\multirow{2}{*}{Method} & \multicolumn{2}{c}{MSCOCO} & \multicolumn{2}{c}{GQA} & \multicolumn{2}{c}{A - OKVQA} \\
\cmidrule(lr){2-3} \cmidrule(lr){4-5} \cmidrule(lr){6-7}
 & acc & F1 & acc & F1 & acc & F1 \\
\midrule
% \scalebox{0.8}{$\begin{array}{c} \text{Logits} \\ \text{Adjustment} \end{array}$} &  \textbf{84.9} & \textbf{85.3} & 79.4 & 82.4 & 80.3 & 82.9  \\
% \scalebox{0.8}{$\begin{array}{c} \text{Bayesian} \\ \text{Rectification} \end{array}$} & 84.8 & 84.8 & \textbf{83.0}	& \textbf{84.0} & \textbf{83.1} & \textbf{84.2} \\

\scalebox{0.9}{Logit Adjustment}  & 84.8 & 84.7 & 79.6 & 80.3 & 81.0 & 83.5 \\
\scalebox{0.9}{Bayesian Rectification} & \textbf{84.8} & \textbf{84.8} & \textbf{83.0}	& \textbf{84.0} & \textbf{83.1} & \textbf{84.2} \\

\bottomrule
\end{tabular}
\label{ablation:bayes_rec}
\end{table}

\textbf{Effect of collapse-aware early stopping.} We validate the effect of our early stopping strategy by comparing it with a baseline method that linearly scales the  $\text{logit}_{\text{eos}}$ according to the generated text length. For a fair comparison, we ensure they have the same recall score. The results are shown in Table \ref{ablation:stop_design}. Our design performs better on both $\text{CHAIR}_\text{S}$ and $\text{CHAIR}_\text{I}$ matrices, verifying the effectiveness of our collapse-aware early stopping approaches.

\setlength{\tabcolsep}{3.5pt}
\begin{table}[h]
\centering
\caption{Effect of collapse-aware early stopping.}
%For a fair comparison, we ensure they have the same recall score.}
\begin{tabular}{lcccc}
\toprule
Method & $\text{CHAIR}_\text{S}\downarrow$ & $\text{CHAIR}_\text{I}\downarrow$ & Recall  & Length  \\
\midrule
Linear  Scaling  & 43.6 & 13.1 & 78.9 & 90.9  \\
Ours            & \textbf{40.2} & \textbf{12.9} & 78.9 & 89.0 \\
\bottomrule
\end{tabular}
\label{ablation:stop_design}
\end{table}

\textbf{Computation efficiency.}
\label{efficiency}
To quantitatively analyze the computation efficiency of EVRB, we compare the inference time of our method, vanilla model, and the previous typical hallucination mitigation method ICD \cite{wang2024mitigating}, and VCD \cite{leng2024mitigating}. Table \ref{time} shows the whole dataset inference time (seconds) under the POPE benchmark, MSCOCO random setting on A100. The inference speed of our method is comparable to the vanilla model, significantly faster than \cite{wang2024mitigating}, and VCD \cite{leng2024mitigating}.

\setlength{\tabcolsep}{20pt}
\begin{table}[h]
\centering
\caption{The whole dataset inference time. }
\begin{tabular}{ll}  
\toprule
Method &  Time \\
\midrule
Vanilla\cite{liu2024improved} & 614s \\
ICD\cite{wang2024mitigating}   & 1151s \\
VCD\cite{leng2024mitigating}     & 1170s \\
EVRB    & 715s \\
\bottomrule
\end{tabular}
\label{time}
\end{table}

\textbf{Generalization to other LVLM.}
We apply EVRB to other LVLMs of different types and scales to further validate the universality of our method.
To verify the universality of EVRB on different types of LVLMs, we employ EVRB on Qwen2.5-vl-7b \cite{bai2025qwen2}, one of the most advantageous LVLMs based on LLM backbone Qwen2.5. It is totally different from LLaVA, which is based on LLM backbone LLaMA. 
For validating scalability across LVLMs, we test EVRB on LLaVA1.5-13b, a larger-scale variant of the LLaVA series.
We conduct experiments on POPE(COCO, GQA, AOK-VQA), CHAIR, and MME benchmarks. All the values of hyperparameters remain unchanged. The results are shown in the Table \ref{more_model}. Compared with other methods, our approach mitigates model hallucinations, demonstrating consistent performance improvement for both discriminative and generative tasks. Also, EVRB improves the performance on MME of both models, showing a comprehensive ability improvement.

\section{CONCLUSION}
In this paper, we address the hallucination problem in LVLMs by comprehensively enhancing the visual reliance in text generation through a Bayesian lens. Our analysis reveals that the decrement of visual reliance during text generation stems from three factors: redundant visual condition, biased prior distribution, and posterior collapse. To tackle these issues, we proposed a training-free framework EVRB. Specifically, we enhance the visual reliance by applying redundant visual token pruning, language prior rectification, and collapse-aware early stopping strategies. Extensive experiments demonstrate that our method performs favorably against existing approaches in mitigating the hallucination issue of LVLM.

\section*{Acknowledgments}
This project is supported by the National Natural Science Foundation of China under Grant 92370114.

{
    \small
    \bibliographystyle{plain}
    \bibliography{main}

\begin{thebibliography}{10}

\bibitem{achiam2023gpt}
Josh Achiam, Steven Adler, Sandhini Agarwal, Lama Ahmad, Ilge Akkaya, Florencia~Leoni Aleman, Diogo Almeida, Janko Altenschmidt, Sam Altman, Shyamal Anadkat, et~al.
\newblock Gpt-4 technical report.
\newblock {\em arXiv preprint arXiv:2303.08774}, 2023.

\bibitem{bai2023qwen}
Jinze Bai, Shuai Bai, Yunfei Chu, Zeyu Cui, Kai Dang, Xiaodong Deng, Yang Fan, Wenbin Ge, Yu~Han, Fei Huang, et~al.
\newblock Qwen technical report.
\newblock {\em arXiv preprint arXiv:2309.16609}, 2023.

\bibitem{bai2025qwen2}
Shuai Bai, Keqin Chen, Xuejing Liu, Jialin Wang, Wenbin Ge, Sibo Song, Kai Dang, Peng Wang, Shijie Wang, Jun Tang, et~al.
\newblock Qwen2. 5-vl technical report.
\newblock {\em arXiv preprint arXiv:2502.13923}, 2025.

\bibitem{bousselham2024grounding}
Walid Bousselham, Felix Petersen, Vittorio Ferrari, and Hilde Kuehne.
\newblock Grounding everything: Emerging localization properties in vision-language transformers.
\newblock In {\em Proceedings of the IEEE/CVF Conference on Computer Vision and Pattern Recognition}, pages 3828--3837, 2024.

\bibitem{brie2023evaluating}
Paul Brie, Nicolas Burny, Arthur Slu{\"y}ters, and Jean Vanderdonckt.
\newblock Evaluating a large language model on searching for gui layouts.
\newblock {\em Proceedings of the ACM on Human-Computer Interaction}, 7(EICS):1--37, 2023.

\bibitem{brown2020language}
Tom Brown, Benjamin Mann, Nick Ryder, Melanie Subbiah, Jared~D Kaplan, Prafulla Dhariwal, Arvind Neelakantan, Pranav Shyam, Girish Sastry, Amanda Askell, et~al.
\newblock Language models are few-shot learners.
\newblock {\em Advances in neural information processing systems}, 33:1877--1901, 2020.

\bibitem{chen2023shikra}
Keqin Chen, Zhao Zhang, Weili Zeng, Richong Zhang, Feng Zhu, and Rui Zhao.
\newblock Shikra: Unleashing multimodal llm's referential dialogue magic.
\newblock {\em arXiv preprint arXiv:2306.15195}, 2023.

\bibitem{chen2024context}
Shiqi Chen, Miao Xiong, Junteng Liu, Zhengxuan Wu, Teng Xiao, Siyang Gao, and Junxian He.
\newblock In-context sharpness as alerts: An inner representation perspective for hallucination mitigation.
\newblock {\em arXiv preprint arXiv:2403.01548}, 2024.

\bibitem{chowdhery2023palm}
Aakanksha Chowdhery, Sharan Narang, Jacob Devlin, Maarten Bosma, Gaurav Mishra, Adam Roberts, Paul Barham, Hyung~Won Chung, Charles Sutton, Sebastian Gehrmann, et~al.
\newblock Palm: Scaling language modeling with pathways.
\newblock {\em Journal of Machine Learning Research}, 24(240):1--113, 2023.

\bibitem{chuang2023dola}
Yung-Sung Chuang, Yujia Xie, Hongyin Luo, Yoon Kim, James Glass, and Pengcheng He.
\newblock Dola: Decoding by contrasting layers improves factuality in large language models.
\newblock {\em arXiv preprint arXiv:2309.03883}, 2023.

\bibitem{cui2024survey}
Can Cui, Yunsheng Ma, Xu~Cao, Wenqian Ye, Yang Zhou, Kaizhao Liang, Jintai Chen, Juanwu Lu, Zichong Yang, Kuei-Da Liao, et~al.
\newblock A survey on multimodal large language models for autonomous driving.
\newblock In {\em Proceedings of the IEEE/CVF Winter Conference on Applications of Computer Vision}, pages 958--979, 2024.

\bibitem{dang2024sadl}
Long~Hoang Dang, Thao~Minh Le, Vuong Le, Tu~Minh Phuong, and Truyen Tran.
\newblock Sadl: An effective in-context learning method for compositional visual qa.
\newblock {\em arXiv preprint arXiv:2407.01983}, 2024.

\bibitem{fu2306mme}
Chaoyou Fu, Peixian Chen, Yunhang Shen, Yulei Qin, Mengdan Zhang, Xu~Lin, Jinrui Yang, Xiawu Zheng, Ke~Li, Xing Sun, et~al.
\newblock Mme: A comprehensive evaluation benchmark for multimodal large language models, 2024.
\newblock {\em URL https://arxiv. org/abs/2306.13394}, 2.

\bibitem{gilardi2023chatgpt}
Fabrizio Gilardi, Meysam Alizadeh, and Ma{\"e}l Kubli.
\newblock Chatgpt outperforms crowd workers for text-annotation tasks.
\newblock {\em Proceedings of the National Academy of Sciences}, 120(30):e2305016120, 2023.

\bibitem{gunjal2024detecting}
Anisha Gunjal, Jihan Yin, and Erhan Bas.
\newblock Detecting and preventing hallucinations in large vision language models.
\newblock In {\em Proceedings of the AAAI Conference on Artificial Intelligence}, volume~38, pages 18135--18143, 2024.

\bibitem{hossain2019comprehensive}
MD~Zakir Hossain, Ferdous Sohel, Mohd~Fairuz Shiratuddin, and Hamid Laga.
\newblock A comprehensive survey of deep learning for image captioning.
\newblock {\em ACM Computing Surveys (CsUR)}, 51(6):1--36, 2019.

\bibitem{hu2023advancing}
Mingzhe Hu, Shaoyan Pan, Yuheng Li, and Xiaofeng Yang.
\newblock Advancing medical imaging with language models: A journey from n-grams to chatgpt.
\newblock {\em arXiv preprint arXiv:2304.04920}, 2023.

\bibitem{huang2024opera}
Qidong Huang, Xiaoyi Dong, Pan Zhang, Bin Wang, Conghui He, Jiaqi Wang, Dahua Lin, Weiming Zhang, and Nenghai Yu.
\newblock Opera: Alleviating hallucination in multi-modal large language models via over-trust penalty and retrospection-allocation.
\newblock In {\em Proceedings of the IEEE/CVF Conference on Computer Vision and Pattern Recognition}, pages 13418--13427, 2024.

\bibitem{hudson2019gqa}
Drew~A Hudson and Christopher~D Manning.
\newblock Gqa: A new dataset for real-world visual reasoning and compositional question answering.
\newblock In {\em Proceedings of the IEEE/CVF conference on computer vision and pattern recognition}, pages 6700--6709, 2019.

\bibitem{huo2024self}
Fushuo Huo, Wenchao Xu, Zhong Zhang, Haozhao Wang, Zhicheng Chen, and Peilin Zhao.
\newblock Self-introspective decoding: Alleviating hallucinations for large vision-language models.
\newblock {\em arXiv preprint arXiv:2408.02032}, 2024.

\bibitem{ji2023survey}
Ziwei Ji, Nayeon Lee, Rita Frieske, Tiezheng Yu, Dan Su, Yan Xu, Etsuko Ishii, Ye~Jin Bang, Andrea Madotto, and Pascale Fung.
\newblock Survey of hallucination in natural language generation.
\newblock {\em ACM computing surveys}, 55(12):1--38, 2023.

\bibitem{leng2024mitigating}
Sicong Leng, Hang Zhang, Guanzheng Chen, Xin Li, Shijian Lu, Chunyan Miao, and Lidong Bing.
\newblock Mitigating object hallucinations in large vision-language models through visual contrastive decoding.
\newblock In {\em Proceedings of the IEEE/CVF Conference on Computer Vision and Pattern Recognition}, pages 13872--13882, 2024.

\bibitem{li2022contrastive}
Xiang~Lisa Li, Ari Holtzman, Daniel Fried, Percy Liang, Jason Eisner, Tatsunori Hashimoto, Luke Zettlemoyer, and Mike Lewis.
\newblock Contrastive decoding: Open-ended text generation as optimization.
\newblock {\em arXiv preprint arXiv:2210.15097}, 2022.

\bibitem{li2023clip}
Yi~Li, Hualiang Wang, Yiqun Duan, and Xiaomeng Li.
\newblock Clip surgery for better explainability with enhancement in open-vocabulary tasks.
\newblock {\em arXiv e-prints}, pages arXiv--2304, 2023.

\bibitem{li2023evaluating}
Yifan Li, Yifan Du, Kun Zhou, Jinpeng Wang, Wayne~Xin Zhao, and Ji-Rong Wen.
\newblock Evaluating object hallucination in large vision-language models.
\newblock {\em arXiv preprint arXiv:2305.10355}, 2023.

\bibitem{liang2024survey}
Zijing Liang, Yanjie Xu, Yifan Hong, Penghui Shang, Qi~Wang, Qiang Fu, and Ke~Liu.
\newblock A survey of multimodel large language models.
\newblock In {\em Proceedings of the 3rd International Conference on Computer, Artificial Intelligence and Control Engineering}, pages 405--409, 2024.

\bibitem{lin2014microsoft}
Tsung-Yi Lin, Michael Maire, Serge Belongie, James Hays, Pietro Perona, Deva Ramanan, Piotr Doll{\'a}r, and C~Lawrence Zitnick.
\newblock Microsoft coco: Common objects in context.
\newblock In {\em Computer vision--ECCV 2014: 13th European conference, zurich, Switzerland, September 6-12, 2014, proceedings, part v 13}, pages 740--755. Springer, 2014.

\bibitem{liu2023aligning}
Fuxiao Liu, Kevin Lin, Linjie Li, Jianfeng Wang, Yaser Yacoob, and Lijuan Wang.
\newblock Aligning large multi-modal model with robust instruction tuning.
\newblock {\em CoRR}, 2023.

\bibitem{liu2023mitigating}
Fuxiao Liu, Kevin Lin, Linjie Li, Jianfeng Wang, Yaser Yacoob, and Lijuan Wang.
\newblock Mitigating hallucination in large multi-modal models via robust instruction tuning.
\newblock {\em arXiv preprint arXiv:2306.14565}, 2023.

\bibitem{liu2024survey}
Hanchao Liu, Wenyuan Xue, Yifei Chen, Dapeng Chen, Xiutian Zhao, Ke~Wang, Liping Hou, Rongjun Li, and Wei Peng.
\newblock A survey on hallucination in large vision-language models.
\newblock {\em arXiv preprint arXiv:2402.00253}, 2024.

\bibitem{liu2024improved}
Haotian Liu, Chunyuan Li, Yuheng Li, and Yong~Jae Lee.
\newblock Improved baselines with visual instruction tuning.
\newblock In {\em Proceedings of the IEEE/CVF Conference on Computer Vision and Pattern Recognition}, pages 26296--26306, 2024.

\bibitem{liu2023visual}
Haotian Liu, Chunyuan Li, Qingyang Wu, and Yong~Jae Lee.
\newblock Visual instruction tuning.
\newblock {\em Advances in neural information processing systems}, 36:34892--34916, 2023.

\bibitem{lu2024insights}
Taiming Lu, Muhan Gao, Kuai Yu, Adam Byerly, and Daniel Khashabi.
\newblock Insights into llm long-context failures: When transformers know but don't tell.
\newblock {\em arXiv preprint arXiv:2406.14673}, 2024.

\bibitem{radford2021learning}
Alec Radford, Jong~Wook Kim, Chris Hallacy, Aditya Ramesh, Gabriel Goh, Sandhini Agarwal, Girish Sastry, Amanda Askell, Pamela Mishkin, Jack Clark, et~al.
\newblock Learning transferable visual models from natural language supervision.
\newblock In {\em International conference on machine learning}, pages 8748--8763. PmLR, 2021.

\bibitem{raffel2020exploring}
Colin Raffel, Noam Shazeer, Adam Roberts, Katherine Lee, Sharan Narang, Michael Matena, Yanqi Zhou, Wei Li, and Peter~J Liu.
\newblock Exploring the limits of transfer learning with a unified text-to-text transformer.
\newblock {\em Journal of machine learning research}, 21(140):1--67, 2020.

\bibitem{rohrbach2018object}
Anna Rohrbach, Lisa~Anne Hendricks, Kaylee Burns, Trevor Darrell, and Kate Saenko.
\newblock Object hallucination in image captioning.
\newblock {\em arXiv preprint arXiv:1809.02156}, 2018.

\bibitem{schwenk2022okvqa}
Dustin Schwenk, Apoorv Khandelwal, Christopher Clark, Kenneth Marino, and Roozbeh Mottaghi.
\newblock A-okvqa: A benchmark for visual question answering using world knowledge.
\newblock In {\em European conference on computer vision}, pages 146--162. Springer, 2022.

\bibitem{sun2023aligning}
Zhiqing Sun, Sheng Shen, Shengcao Cao, Haotian Liu, Chunyuan Li, Yikang Shen, Chuang Gan, Liang-Yan Gui, Yu-Xiong Wang, Yiming Yang, et~al.
\newblock Aligning large multimodal models with factually augmented rlhf.
\newblock {\em arXiv preprint arXiv:2309.14525}, 2023.

\bibitem{taori2023stanford}
Rohan Taori, Ishaan Gulrajani, Tianyi Zhang, Yann Dubois, Xuechen Li, Carlos Guestrin, Percy Liang, and Tatsunori~B Hashimoto.
\newblock Stanford alpaca: An instruction-following llama model, 2023.

\bibitem{touvron2023llama}
Hugo Touvron, Thibaut Lavril, Gautier Izacard, Xavier Martinet, Marie-Anne Lachaux, Timoth{\'e}e Lacroix, Baptiste Rozi{\`e}re, Naman Goyal, Eric Hambro, Faisal Azhar, et~al.
\newblock Llama: Open and efficient foundation language models.
\newblock {\em arXiv preprint arXiv:2302.13971}, 2023.

\bibitem{wang2024mllm}
Chenxi Wang, Xiang Chen, Ningyu Zhang, Bozhong Tian, Haoming Xu, Shumin Deng, and Huajun Chen.
\newblock Mllm can see? dynamic correction decoding for hallucination mitigation.
\newblock {\em arXiv preprint arXiv:2410.11779}, 2024.

\bibitem{wang2024sclip}
Feng Wang, Jieru Mei, and Alan Yuille.
\newblock Sclip: Rethinking self-attention for dense vision-language inference.
\newblock In {\em European Conference on Computer Vision}, pages 315--332. Springer, 2024.

\bibitem{wang2024surgical}
Guankun Wang, Long Bai, Wan~Jun Nah, Jie Wang, Zhaoxi Zhang, Zhen Chen, Jinlin Wu, Mobarakol Islam, Hongbin Liu, and Hongliang Ren.
\newblock Surgical-lvlm: Learning to adapt large vision-language model for grounded visual question answering in robotic surgery.
\newblock {\em arXiv preprint arXiv:2405.10948}, 2024.

\bibitem{wang2024qwen2}
Peng Wang, Shuai Bai, Sinan Tan, Shijie Wang, Zhihao Fan, Jinze Bai, Keqin Chen, Xuejing Liu, Jialin Wang, Wenbin Ge, et~al.
\newblock Qwen2-vl: Enhancing vision-language model's perception of the world at any resolution.
\newblock {\em arXiv preprint arXiv:2409.12191}, 2024.

\bibitem{wang2023chatcad}
Sheng Wang, Zihao Zhao, Xi~Ouyang, Qian Wang, and Dinggang Shen.
\newblock Chatcad: Interactive computer-aided diagnosis on medical image using large language models.
\newblock {\em arXiv preprint arXiv:2302.07257}, 2023.

\bibitem{wang2023caption}
Teng Wang, Jinrui Zhang, Junjie Fei, Hao Zheng, Yunlong Tang, Zhe Li, Mingqi Gao, and Shanshan Zhao.
\newblock Caption anything: Interactive image description with diverse multimodal controls.
\newblock {\em arXiv preprint arXiv:2305.02677}, 2023.

\bibitem{wang2023drivemlm}
Wenhai Wang, Jiangwei Xie, ChuanYang Hu, Haoming Zou, Jianan Fan, Wenwen Tong, Yang Wen, Silei Wu, Hanming Deng, Zhiqi Li, et~al.
\newblock Drivemlm: Aligning multi-modal large language models with behavioral planning states for autonomous driving.
\newblock {\em arXiv preprint arXiv:2312.09245}, 2023.

\bibitem{wang2024mitigating}
Xintong Wang, Jingheng Pan, Liang Ding, and Chris Biemann.
\newblock Mitigating hallucinations in large vision-language models with instruction contrastive decoding.
\newblock {\em arXiv preprint arXiv:2403.18715}, 2024.

\bibitem{yao2024minicpm}
Yuan Yao, Tianyu Yu, Ao~Zhang, Chongyi Wang, Junbo Cui, Hongji Zhu, Tianchi Cai, Haoyu Li, Weilin Zhao, Zhihui He, et~al.
\newblock Minicpm-v: A gpt-4v level mllm on your phone.
\newblock {\em arXiv preprint arXiv:2408.01800}, 2024.

\bibitem{yin2024woodpecker}
Shukang Yin, Chaoyou Fu, Sirui Zhao, Tong Xu, Hao Wang, Dianbo Sui, Yunhang Shen, Ke~Li, Xing Sun, and Enhong Chen.
\newblock Woodpecker: Hallucination correction for multimodal large language models.
\newblock {\em Science China Information Sciences}, 67(12):220105, 2024.

\bibitem{yue2024less}
Zihao Yue, Liang Zhang, and Qin Jin.
\newblock Less is more: Mitigating multimodal hallucination from an eos decision perspective.
\newblock {\em arXiv preprint arXiv:2402.14545}, 2024.

\bibitem{zhang2023huatuogpt}
Hongbo Zhang, Junying Chen, Feng Jiang, Fei Yu, Zhihong Chen, Jianquan Li, Guiming Chen, Xiangbo Wu, Zhiyi Zhang, Qingying Xiao, et~al.
\newblock Huatuogpt, towards taming language model to be a doctor.
\newblock {\em arXiv preprint arXiv:2305.15075}, 2023.

\bibitem{zhao2023beyond}
Zhiyuan Zhao, Bin Wang, Linke Ouyang, Xiaoyi Dong, Jiaqi Wang, and Conghui He.
\newblock Beyond hallucinations: Enhancing lvlms through hallucination-aware direct preference optimization.
\newblock {\em arXiv preprint arXiv:2311.16839}, 2023.

\bibitem{zhou2024mitigating}
Guanyu Zhou, Yibo Yan, Xin Zou, Kun Wang, Aiwei Liu, and Xuming Hu.
\newblock Mitigating modality prior-induced hallucinations in multimodal large language models via deciphering attention causality.
\newblock {\em arXiv preprint arXiv:2410.04780}, 2024.

\bibitem{zhu2023minigpt}
Deyao Zhu, Jun Chen, Xiaoqian Shen, Xiang Li, and Mohamed Elhoseiny.
\newblock Minigpt-4: Enhancing vision-language understanding with advanced large language models.
\newblock {\em arXiv preprint arXiv:2304.10592}, 2023.

\end{thebibliography}
}

\clearpage
\section*{APPENDIX}

\subsection{Validations on more models}
 To further validate the generality of our approach, we conducted additional experiments on another LVLM, Shikra \cite{chen2023shikra}, using both the POPE \cite{li2023evaluating} and CHAIR \cite{rohrbach2018object} benchmarks. Table \ref{ablation:model} shows the overall results, and, specifically, the POPE scores are averaged under the three sub-tasks of three datasets. Compared with other methods, our approach mitigates model hallucinations and demonstrates consistent performance improvement for both discriminative and generative tasks.

\setlength{\tabcolsep}{15pt}
\begin{table*}[t]
\centering
\begin{tabular}{lcccccc}
\toprule
\multirow{2}{*}{Method} & \multicolumn{4}{c}{POPE}  & \multicolumn{2}{c}{CHAIR} \\
\cmidrule(lr){2-5} \cmidrule(lr){6-7}
 & Accuracy & Precision & Recall & F1 &$\text{CHAIR}_\text{S}\downarrow$ & $\text{CHAIR}_\text{I}\downarrow$\\
\midrule
Vanilla \cite{chen2023shikra} & 70.8 & 65.0 & \textbf{90.8}  & 75.7  & 56.6 & 16.6\\
VCD \cite{leng2024mitigating}    & 68.3 & 61.8 & 90.6  & 73.5  & 55.0 & 15.4\\
SID \cite{huo2024self}    & 64.3 & 59.1 & 90.1 & 71.6  & 54.2 & 14.3\\
Ours    & \textbf{72.4} & \textbf{66.8} & 89.9 & \textbf{76.7}  & \textbf{44.3} & \textbf{14.1} \\
\bottomrule
\end{tabular}
\caption{POPE and CHAIR hallucination evaluation results based on Shikra.}
\label{ablation:model}
\end{table*}

\subsection{Hyper-parameters robustness}

We conduct ablation studies on all hyper-parameters, including scaling ratio for early stop $\lambda$, threshold for visual-relevant tokens selection $\delta$, visual token pruning threshold $\tau$, and probability truncation ratio $\mu$. The results are presented in the following Table \ref{hyperparam}. We can observe from the table that our method remains relatively robust to the choices of hyper-parameters within a certain range. Besides, the $\lambda$ and $\delta$ are selected to strike a good balance between hallucination mitigation (CHAIRs/i) and generation completeness (Recall).
It is worth noting that the optimal value of $\tau$ is also the valley of the bimodal distribution of the statistical result, further justifying its selection. 

Moreover, our hyper-parameters choices can generalize to different LVLMs, e.g., Qwen2.5-vl-7b\cite{bai2025qwen2} and LLaVA-1.5-13b\cite{liu2024improved}. With the same hyperparameters, the superior hallucination mitigation results are obtained. 

\setlength{\tabcolsep}{2.5pt}
\begin{table*}[h]
\centering
\caption{Robustness analysis on all hyper-parameters, including scaling ratio for early stop $\lambda$, threshold for visual-relevant tokens selection $\delta$, visual token pruning threshold $\tau$, and probability truncation ratio $\mu$.}
\renewcommand{\arraystretch}{1.2}

\begin{tabular}{cccccccccccccc}
\toprule

$\lambda$ & {CHAIRs$\downarrow$} & {CHAIRi$\downarrow$} & {Recall} & 
$\delta$  & {CHAIRs$\downarrow$} & {CHAIRi$\downarrow$} & {Recall} & 
$\tau$    & {Accuracy}           & {F1-score}         & 
$\mu$     & {Accuracy}           & {F1-score}         \\
\cmidrule(lr){1-4}  \cmidrule(lr){5-8}  \cmidrule(lr){9-11}  \cmidrule(lr){12-14}

1.44 & 40.2 & 12.9 & 78.6 & 0.044 & 39.0 & 12.7 & 77.7 & 0.44 & 0.825 & 0.834 & 0.04 & 0.846 & 0.849 \\
1.47 & 40.0 & 12.8 & 78.5 & 0.047 & 39.4 & 12.8 & 78.0 & 0.47 & 0.831 & 0.839 & 0.07 & 0.846 & 0.849 \\
1.50 & 39.8 & 12.8 & 78.5 & 0.050 & 39.8 & 12.8 & 78.5 & 0.50 & 0.848 & 0.849 & 0.10 & 0.848 & 0.849 \\
1.53 & 39.8 & 12.8 & 78.5 & 0.053 & 40.6 & 12.9 & 78.8 & 0.53 & 0.84  & 0.845 & 0.13 & 0.838 & 0.844 \\
1.56 & 39.6 & 12.8 & 78.4 & 0.056 & 41.0 & 13.1 & 78.7 & 0.56 & 0.840 & 0.846 & 0.16 & 0.838 & 0.845 \\
\bottomrule
\end{tabular}
\label{hyperparam}
\end{table*}

\subsection{More samples for text-to-visual attention maps}
\begin{figure*}[h]
  \centering
  \includegraphics[width=\linewidth]{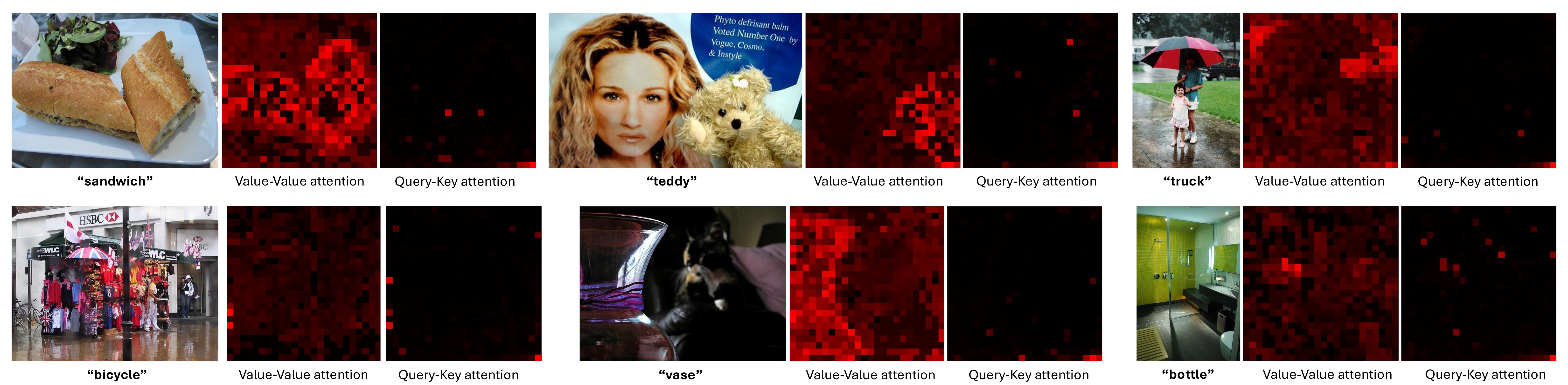} 
  \caption{Additionaly text-to-visual attention maps samples} 
  \label{vv_maps}
\end{figure*}

In Section \ref{early stopping}, we propose the collapse-aware early stopping strategy to adjust the termination probability, and the value-value attention map was leveraged to select the visual-relevant textual tokens. We provide more text-to-visual attention maps in \ref{vv_maps} to further display the semantic coherence between the value features of visual-relevant text and visual tokens. 

\subsection{Case study}
We qualitatively show the hallucination issues and the hallucination mitigation effects of different methods. 

\begin{figure*}[h]
    \raggedright
  \subfloat[]{
      \includegraphics[width=0.777\textwidth]{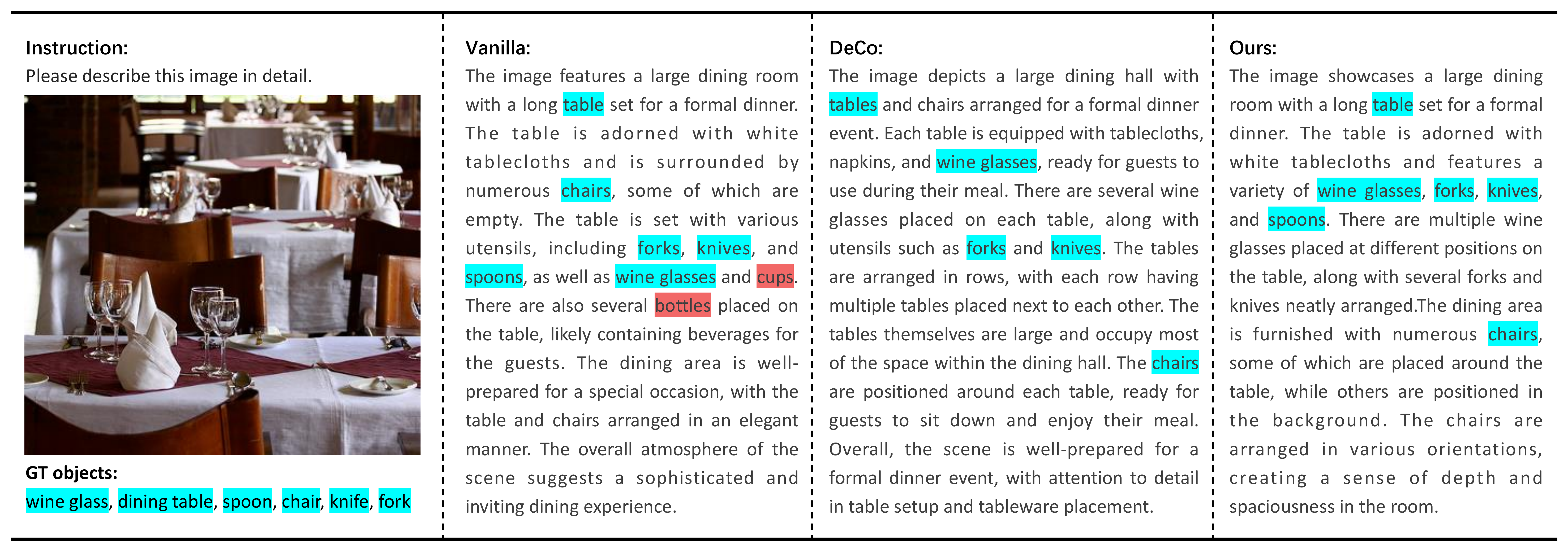}
      \label{case_chair}
  }
  \subfloat[]{
      \includegraphics[width=0.223\textwidth]{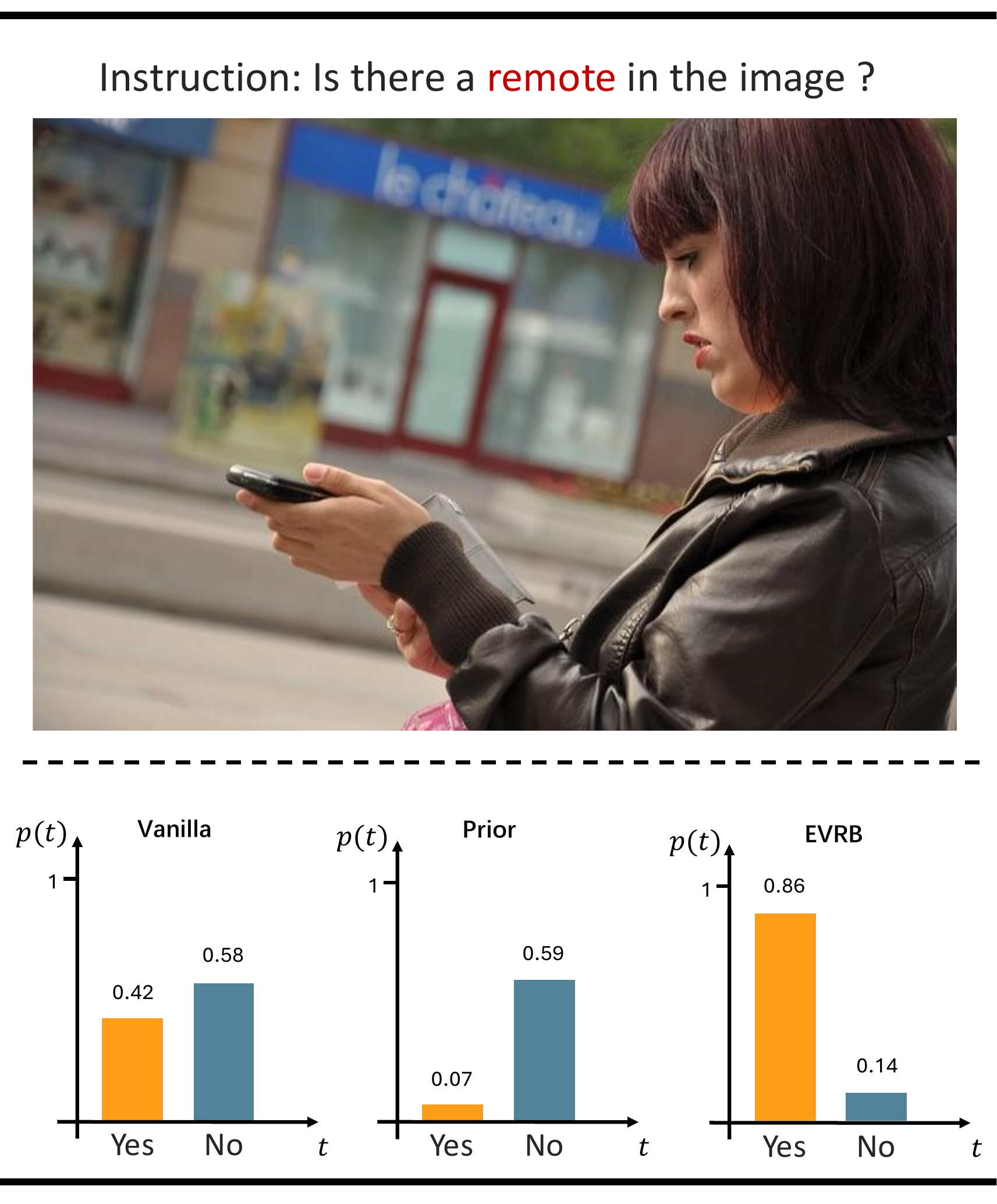}
      \label{case_pope}
      
  }
    \caption{Case study for the hallucination mitigation in generative (a) and discriminative (b) tasks }
    \label{case_study}
\end{figure*}

Fig. \ref{case_chair} shows the hallucination issues that happened in the generative task. In this scenario, the vanilla LVLM tends to describe some objects that do not exist in the input image, \emph{e.g.,} cups, bottles, \emph{etc.}, exhibiting a hallucination issue. Both DeCo and our method can mitigate the hallucination issue in this case. However, DeCo fails to describe all the ground-truth objects, \emph{e.g.,} overlooking the ``spoon'' in the image. In contrast, our method describes all the ground-truth objects, exhibiting superior generation ability and hallucination mitigation ability. 

In the discrimination task, we can observe an obvious language prior bias that can lead to Hallucination. As Fig. \ref{case_pope} shows, the vanilla model fails to correctly judge the existence of remote in the image. And if we apply a pure language prior inference with the input that excludes the image, we can find an obvious bias to answer ``No''. The failure of the vanilla model can contribute to this language prior bias, and our proposed method EVRB can effectively eliminate this bias and mitigate the hallucination. 

\end{document}